%% file: paper.tex
\documentclass{article}
\PassOptionsToPackage{table}{xcolor}
\usepackage{iclr2027_conference,times}
\usepackage[T1]{fontenc}
\usepackage[utf8]{inputenc}
\usepackage{amsmath,amssymb,graphicx,booktabs,tabularx,longtable,array}
\usepackage[table]{xcolor}
\usepackage{etoolbox,placeins}
\AtBeginEnvironment{table}{\setlength{\belowcaptionskip}{10pt}}
\makeatletter
\def\LT@makecaption#1#2#3{%
  \LT@mcol\LT@cols {@{}l@{}}{\parbox[t]{\linewidth}{%
    \normalfont\normalsize
    \sbox\@tempboxa{#1{#2: }#3}%
    \ifdim\wd\@tempboxa>\hsize
      #1{#2: }#3%
    \else
      \hbox to\hsize{\hfil\box\@tempboxa\hfil}%
    \fi
    \endgraf\vskip 10pt}}}
\makeatother

\usepackage{hyperref,url}
\hypersetup{
  colorlinks=true,
  linkcolor=blue,
  citecolor=blue,
  urlcolor=blue,
  pdftitle={ResearchFlow-Eval: Tracing the Execution Gap in End-to-End Multi-Claim Autoresearch},
  pdfauthor={Peng Jiang; Dongyang Li; Quanying Liu; Xiaoxuan Jia},
  pdfsubject={Autoresearch systems evaluation},
  pdfkeywords={autoresearch, AI agents, evaluation, multi-claim workflows, reproducibility}
}
\newcommand{\method}{ResearchFlow-Eval}

\newcommand{\artifacturl}{https://github.com/Jp-17/ResearchFlow-Eval}
\title{ResearchFlow-Eval: Tracing the Execution Gap in End-to-End Multi-Claim Autoresearch}
\author{%
Peng Jiang\textsuperscript{1,2,3} \quad
Dongyang Li\textsuperscript{1,2,3} \quad
Quanying Liu\textsuperscript{4,5}\thanks{Corresponding authors: \texttt{liuqy@sustech.edu.cn}, \texttt{jiaxx@mail.tsinghua.edu.cn}.} \quad
Xiaoxuan Jia\textsuperscript{1,2,3}\footnotemark[1]\\
\normalfont\textsuperscript{1}Center for Life Sciences and Artificial Intelligence, School of Life Sciences,\\
Tsinghua University, Beijing, China\\
\textsuperscript{2}IDG/McGovern Institute for Brain Research, Tsinghua University, Beijing, China\\
\textsuperscript{3}Tsinghua-Peking Center for Life Sciences, Tsinghua University, Beijing, China\\
\textsuperscript{4}Department of Biomedical Engineering,\\
Southern University of Science and Technology, Shenzhen, China\\
\textsuperscript{5}Omni-Intelligence, Shenzhen, China
}
\iclrfinalcopy
\begin{document}
\maketitle
\enlargethispage{\baselineskip}
\fancyhead{}
\renewcommand{\headrulewidth}{0pt}
\begin{abstract}
Autoresearch systems increasingly undertake entire research projects. In practice, scientific research requires systems to pursue multiple interdependent claims through a continuous, multi-stage workflow and revise experiments as evidence accumulates. Existing evaluations of autoresearch systems, however, typically score isolated tasks or final artifacts, leaving these process-level capabilities largely unmeasured. We therefore introduce ResearchFlow-Eval, a fine-grained framework for assessing continuous scientific execution and iteration at both the stage and claim levels. Each evaluation is grounded in a published anchor paper, from which we derive stage-specific rubrics that decompose each research stage into dimensions and separately scored rubric items. Using these rubrics, we score each research stage and examine experimental execution in detail, focusing on multi-claim completion and revisions across attempts. We apply the evaluation framework to general-purpose agent harnesses, skill-augmented harnesses, and systems with research-specific orchestration. Averaged over two runs per system, each scored by two LLM judges, experimental execution lags behind design and planning for every evaluated system. Within execution, systems struggle to produce high-quality results and meet claim requirements, and repeated attempts rarely yield measurable improvements. Moreover, neither skill augmentation nor research-specific orchestration reliably yields higher scores. ResearchFlow-Eval thus extends autoresearch evaluation beyond isolated, single-objective runs to continuous workflows in which systems pursue multiple claims and refine their work across iterations.
\end{abstract}
\input{sections/introduction}
\raggedbottom
\input{sections/related_work}
\input{sections/method}
\input{sections/setup}
\input{sections/results}
\begin{samepage}
\input{sections/discussion}
\par
\end{samepage}
\section*{Ethics statement}
This work evaluates research artifacts produced by autoresearch systems. The dataset used in this work is publicly released and is used under its original access terms. This work collects no new data from human participants or animals and raises no additional ethical concerns.
\section*{Reproducibility statement}
Section~\ref{sec:task-construction} describes how the task and its required claims are derived from the anchor paper and frozen before any system is run. Section~\ref{sec:scoring} specifies the rubrics, the score aggregation, and the pinned LLM judge, and Section~\ref{sec:setup} specifies the evaluated systems, the run environment, and the resource budgets. The task package and evaluation code are available at \href{\artifacturl}{\nolinkurl{\artifacturl}}.
\section*{AI use statement}
Large language models are part of the object of study, because the evaluated autoresearch systems are LLM agents. In research execution, the authors used generative AI tools for limited assistance with setting up the evaluation system, literature search and summarization, figure plotting, and manuscript refinement. All AI-assisted content was reviewed by the authors, who take responsibility for the final work.
\bibliography{references}
\bibliographystyle{iclr2027_conference}
\clearpage
\flushbottom
\appendix
\input{sections/appendices}
\end{document}

%% file: sections/introduction.tex
\section{Introduction}
Autoresearch systems increasingly undertake entire research projects, from literature search and idea formulation to coding, experimentation, analysis, and manuscript writing \citep{lu_towards_2026,schmidgall_agent_2025,yamada_ai_2025,weng_deepscientist_2025,liu_autoresearchclaw_2026}. A research project, however, is not a single task. It advances several interdependent claims through a continuous, multi-stage workflow in which each stage builds on the outputs of earlier stages, and its experiments must be revised as evidence accumulates. Existing benchmarks each cover part of this workflow. They score paper replication \citep{starace_paperbench_2025}, single experiments derived from published papers \citep{kon_exp-bench_2025}, iterative experimentation \citep{garikaparthi_researchgym_2026,zou_fml-bench_2025}, individual research stages \citep{chen_mlr-bench_2025,cui_arac_2026}, or the outputs of a complete project \citep{zhou_asibench_2026,xu_researchclawbench_2026}. Their scores, however, typically attach to an isolated task, a final artifact, or a single optimization objective, rather than following multiple required claims through one continuous workflow. Such scores reveal how far a system got, but not whether its plans were carried through to supporting evidence. This matters because autoresearch systems can produce sound proposals and fluent manuscripts while the experiments behind their conclusions remain unfinished or unverified \citep{trehan_why_2026,chen_mlr-bench_2025,zhang_how_2026}. We refer to this disparity between planning a multi-claim research project and carrying each of its claims through to supporting evidence as the \emph{execution gap}.

\begin{figure}[t]
\setlength{\abovecaptionskip}{7pt}
\centering\includegraphics[width=0.82\linewidth]{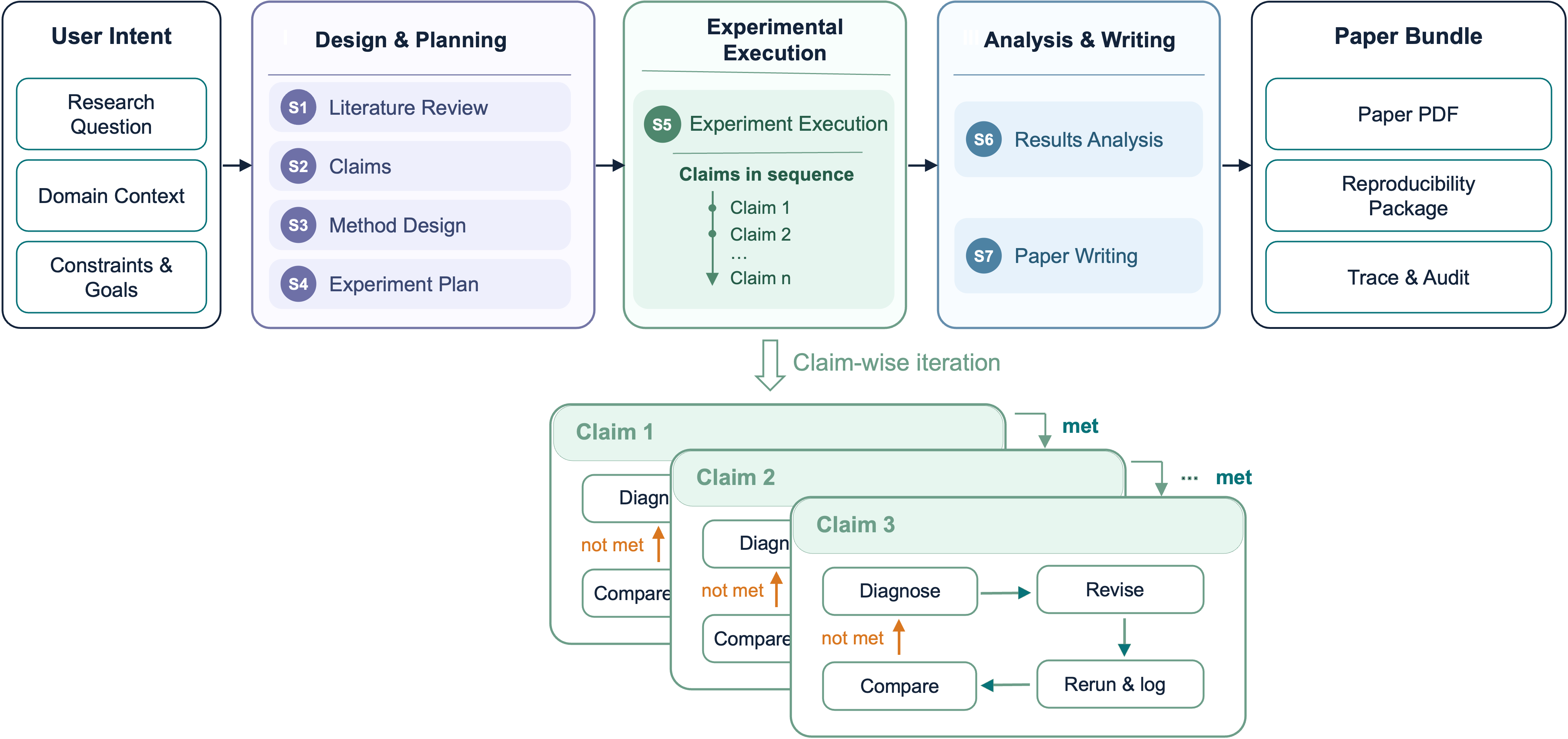}
\caption{Overview of the multi-claim research workflow. Design and planning (S1--S4) precede experimental execution (S5) and analysis and writing (S6--S7). The inset shows claim-wise iteration, in which an unmet criterion returns the claim to diagnosis and a met one moves the system to the next claim.}
\label{fig:framework}
\end{figure}

{\looseness=-1
Measuring the execution gap requires following each required claim through the workflow and tracking how its evidence changes across attempts. To this end, we introduce \textbf{\method{}}, a fine-grained framework for evaluating continuous scientific execution and iteration at both the stage and claim levels. Each evaluation is grounded in a published anchor paper, from which we derive stage-specific rubrics for seven stages (Figure~\ref{fig:framework}) and an explicit success criterion for each required claim. Because every rubric item is tied to evidence in the anchor paper and in the system's own artifacts, each score can be inspected and recomputed. Pairing end-to-end runs with stagewise runs, which start each stage from paper-derived inputs, separates a stage's own failures from errors inherited from upstream. Within experimental execution, claim-level verdicts show how many required claims a system establishes and whether its revisions across attempts move them toward their criteria.\par}

We evaluate general-purpose agent harnesses (Codex \citep{openai_codex}, Claude Code \citep{anthropic_claude_code}, and Kimi Code \citep{moonshot_kimi_code}), skill-augmented harnesses (ARIS \citep{yang_aris_2026} and Orchestra \citep{orchestra_ai_research_skills}), and systems with research-specific orchestration (AutoResearchClaw \citep{liu_autoresearchclaw_2026} and DeepScientist \citep{weng_deepscientist_2025}). Averaged over two runs per system, each scored by two LLM judges, experimental execution scores below design and planning for every system. Stagewise runs of the general-purpose harnesses raise the overall score in every run under both judges, suggesting that part of this deficit arises upstream of execution. The deficit is sharpest when each required claim is checked against its success criterion. In the first run, the evaluated systems report success on as many as eight required claims, yet at most four meet their criteria, and repeated attempts rarely turn an unmet claim into a met one. Their self-reported progress is thus a poor proxy for scientific progress \citep{trehan_why_2026,chen_mlr-bench_2025}. Adding research-specific structure does not narrow the gap either.

\begin{samepage}
We summarize our contributions as follows:
\begin{itemize}
\setlength{\itemsep}{1pt}
\setlength{\parskip}{0pt}
\setlength{\parsep}{0pt}
\setlength{\topsep}{3pt}
    \item \textbf{Evaluating continuous multi-claim scientific workflows.} We evaluate autoresearch systems on an entire research project rather than an isolated task. Our multi-level assessment measures how far each required claim is completed and diagnoses how systems iterate.
    \item \textbf{Fine-grained scoring across stages and dimensions.} We derive stage-specific rubrics from an anchor paper and decompose each stage into dimensions and separately scored rubric items. Every score is tied to evidence that can be inspected and recomputed.
    \item \textbf{Effects of skill augmentation and research orchestration.} We compare general-purpose harnesses with skill-augmented harnesses and research orchestration systems on the same backbone model, over two runs and two judges. Neither reliably scores higher than the bare harness, and neither closes the execution gap.
\end{itemize}
\end{samepage}

%% file: sections/related_work.tex
\section{Related work}\label{sec:related}
\subsection{Autonomous scientific research systems}
In autonomous research systems, language-model agents carry out increasingly complete research workflows, and these systems differ mainly in how much of the research process they organize explicitly. General-purpose agent harnesses such as Codex, Claude Code, and Kimi Code run a generic tool loop that reads files, edits code, and executes commands \citep{openai_codex,anthropic_claude_code,moonshot_kimi_code}. They are not designed for research, yet they are increasingly run directly on research tasks \citep{zhang_how_2026,xu_researchclawbench_2026,garikaparthi_researchgym_2026}. Skill-augmented harnesses keep this loop and add reusable research skills, such as a cross-model reviewer in ARIS and a research-engineering skill library in Orchestra \citep{yang_aris_2026,orchestra_ai_research_skills}. Research-orchestration systems instead organize the research lifecycle with their own workflow, state, and agent coordination. The AI Scientist, Agent Laboratory, and AI-Researcher chain ideation, experiments, and writing into end-to-end pipelines \citep{lu_towards_2026,yamada_ai_2025,schmidgall_agent_2025,tang_ai-researcher_2025}. AutoResearchClaw adds multi-agent debate and failure recovery \citep{liu_autoresearchclaw_2026}, and DeepScientist runs an iterative discovery loop with a cumulative findings memory \citep{weng_deepscientist_2025}. These systems are mostly judged by the papers they produce, which can overstate progress when the underlying experiments are weak \citep{zhang_how_2026}. Comparing them therefore requires an evaluation that observes how each stage and each claim is executed.

\subsection{Benchmarks for autonomous research agents}
Evaluation of research agents has progressed from isolated execution tasks toward longer research workflows (Table~\ref{tab:benchmarks}).
\begin{table}[!t]
\setlength{\tabcolsep}{5pt}
\caption{Comparison with related autoresearch benchmarks. The research-part columns mark which stages a benchmark evaluates. The evaluation-feature columns mark whether it scores a full end-to-end run and multiple claims per run, whether its rubric is derived from a paper, and whether it diagnoses iteration within a task. \textcolor{green!45!black}{$\checkmark$}, \textcolor{orange!80!black}{$\triangle$}, and \textcolor{red!70!black}{$\times$} indicate that a benchmark evaluates an item explicitly, partially, or not at all.}
\label{tab:benchmarks}
\centering\small\renewcommand{\arraystretch}{1.08}\input{tables/tab01}
\end{table}
CORE-Bench and ScienceAgentBench score reproduction and scientific programs \citep{siegel_core-bench_2026,chen_scienceagentbench_2025}, while MLAgentBench, MLE-bench, RE-Bench, and FML-bench measure how far experiments improve a task metric \citep{huang_mlagentbench_2023,chan_mle-bench_2025,wijk_re-bench_2024,zou_fml-bench_2025}. PaperBench, ResearchGym, and FIRE-Bench place execution inside a paper-derived research task and score replication, improvement, or rediscovered findings \citep{starace_paperbench_2025,garikaparthi_researchgym_2026,wang_fire-bench_2026}, while ASI-Bench scores project-level research against reference outputs \citep{zhou_asibench_2026}. MLR-Bench and ARAC score each research stage, and ResearchClawBench scores the report and artifacts of a complete run \citep{chen_mlr-bench_2025,cui_arac_2026,xu_researchclawbench_2026}. Some benchmarks also track performance over a run or diagnose failure patterns such as repeated loops and premature termination \citep{garikaparthi_researchgym_2026,zou_fml-bench_2025,chen_scienceagentbench_2025,starace_paperbench_2025,wang_fire-bench_2026}. EXP-Bench derives experiment tasks from published papers, scores experimental design, implementation, execution, and conclusions, and finds that agents rarely complete an executable experiment even when they handle its parts \citep{kon_exp-bench_2025}. Most of these benchmarks score one stage given reference inputs, one experiment, or the final artifact of a run. \method{} builds on them by following several interdependent claims through one continuous run in which later stages consume the system's own earlier outputs. It derives stage rubrics and claim requirements from an anchor paper, scores every stage, and tracks how the evidence for each claim develops across attempts.

%% file: tables/tab01.tex
\begin{tabularx}{\linewidth}{>{\raggedright\arraybackslash}p{.23\linewidth}*{6}{>{\centering\arraybackslash}X}>{\centering\arraybackslash}p{.115\linewidth}}
\toprule
 & \multicolumn{3}{c}{Research parts} & \multicolumn{4}{c}{Evaluation features} \\
\cmidrule(lr){2-4}\cmidrule(l){5-8}
Benchmark & \shortstack{Design\\plan} & \shortstack{Exp.\\execute} & \shortstack{Analyze\\write} & \shortstack{Full\\E2E} & \shortstack{Multi-\\claim} & \shortstack{Paper\\rubric} & \shortstack{Iteration\\diagnostics} \\ \midrule
\rowcolor{black!2}
Science\allowbreak AgentBench \citep{chen_scienceagentbench_2025} & \textcolor{red!70!black}{$\times$} & \textcolor{orange!80!black}{$\triangle$} & \textcolor{orange!80!black}{$\triangle$} & \textcolor{red!70!black}{$\times$} & \textcolor{red!70!black}{$\times$} & \textcolor{green!45!black}{$\checkmark$} & \textcolor{green!45!black}{$\checkmark$} \\
PaperBench \citep{starace_paperbench_2025} & \textcolor{red!70!black}{$\times$} & \textcolor{green!45!black}{$\checkmark$} & \textcolor{orange!80!black}{$\triangle$} & \textcolor{orange!80!black}{$\triangle$} & \textcolor{green!45!black}{$\checkmark$} & \textcolor{green!45!black}{$\checkmark$} & \textcolor{green!45!black}{$\checkmark$} \\
\rowcolor{black!2}
\leavevmode EXP-Bench \citep{kon_exp-bench_2025} & \textcolor{green!45!black}{$\checkmark$} & \textcolor{green!45!black}{$\checkmark$} & \textcolor{orange!80!black}{$\triangle$} & \textcolor{orange!80!black}{$\triangle$} & \textcolor{red!70!black}{$\times$} & \textcolor{green!45!black}{$\checkmark$} & \textcolor{red!70!black}{$\times$} \\
MLR-Bench \citep{chen_mlr-bench_2025} & \textcolor{green!45!black}{$\checkmark$} & \textcolor{green!45!black}{$\checkmark$} & \textcolor{green!45!black}{$\checkmark$} & \textcolor{green!45!black}{$\checkmark$} & \textcolor{red!70!black}{$\times$} & \textcolor{red!70!black}{$\times$} & \textcolor{orange!80!black}{$\triangle$} \\
\rowcolor{black!2}
\leavevmode ARAC \citep{cui_arac_2026} & \textcolor{green!45!black}{$\checkmark$} & \textcolor{orange!80!black}{$\triangle$} & \textcolor{green!45!black}{$\checkmark$} & \textcolor{red!70!black}{$\times$} & \textcolor{red!70!black}{$\times$} & \textcolor{green!45!black}{$\checkmark$} & \textcolor{red!70!black}{$\times$} \\
ResearchGym \citep{garikaparthi_researchgym_2026} & \textcolor{orange!80!black}{$\triangle$} & \textcolor{green!45!black}{$\checkmark$} & \textcolor{red!70!black}{$\times$} & \textcolor{orange!80!black}{$\triangle$} & \textcolor{orange!80!black}{$\triangle$} & \textcolor{orange!80!black}{$\triangle$} & \textcolor{green!45!black}{$\checkmark$} \\
\rowcolor{black!2}
FML-bench \citep{zou_fml-bench_2025} & \textcolor{red!70!black}{$\times$} & \textcolor{green!45!black}{$\checkmark$} & \textcolor{red!70!black}{$\times$} & \textcolor{orange!80!black}{$\triangle$} & \textcolor{red!70!black}{$\times$} & \textcolor{red!70!black}{$\times$} & \textcolor{green!45!black}{$\checkmark$} \\
FIRE-Bench \citep{wang_fire-bench_2026} & \textcolor{orange!80!black}{$\triangle$} & \textcolor{green!45!black}{$\checkmark$} & \textcolor{orange!80!black}{$\triangle$} & \textcolor{orange!80!black}{$\triangle$} & \textcolor{orange!80!black}{$\triangle$} & \textcolor{green!45!black}{$\checkmark$} & \textcolor{green!45!black}{$\checkmark$} \\
\rowcolor{black!2}
ResearchClaw\allowbreak Bench \citep{xu_researchclawbench_2026} & \textcolor{orange!80!black}{$\triangle$} & \textcolor{green!45!black}{$\checkmark$} & \textcolor{green!45!black}{$\checkmark$} & \textcolor{green!45!black}{$\checkmark$} & \textcolor{orange!80!black}{$\triangle$} & \textcolor{green!45!black}{$\checkmark$} & \textcolor{red!70!black}{$\times$} \\
ASI-Bench \citep{zhou_asibench_2026} & \textcolor{orange!80!black}{$\triangle$} & \textcolor{green!45!black}{$\checkmark$} & \textcolor{orange!80!black}{$\triangle$} & \textcolor{orange!80!black}{$\triangle$} & \textcolor{red!70!black}{$\times$} & \textcolor{red!70!black}{$\times$} & \textcolor{red!70!black}{$\times$} \\
\midrule\noalign{\vskip-\belowrulesep}\rowcolor{violet!10}
\textbf{ResearchFlow-Eval (ours)} & \textcolor{green!45!black}{$\checkmark$} & \textcolor{green!45!black}{$\checkmark$} & \textcolor{green!45!black}{$\checkmark$} & \textcolor{green!45!black}{$\checkmark$} & \textcolor{green!45!black}{$\checkmark$} & \textcolor{green!45!black}{$\checkmark$} & \textcolor{green!45!black}{$\checkmark$} \\
\noalign{\vskip-\aboverulesep}\bottomrule\end{tabularx}

%% file: sections/method.tex
\section{ResearchFlow-Eval}\label{sec:framework}
\method{} evaluates autoresearch systems on research projects derived from published anchor papers. In each task, the evaluated autoresearch system receives a research brief and access to data. It then carries the project through seven research stages, from literature review to paper writing, and submits a required artifact at each stage (Figure~\ref{fig:framework}). We score these artifacts against a paper-derived reference that is withheld from the evaluated system.

\begin{figure}[!htb]
\centering\includegraphics[width=\linewidth]{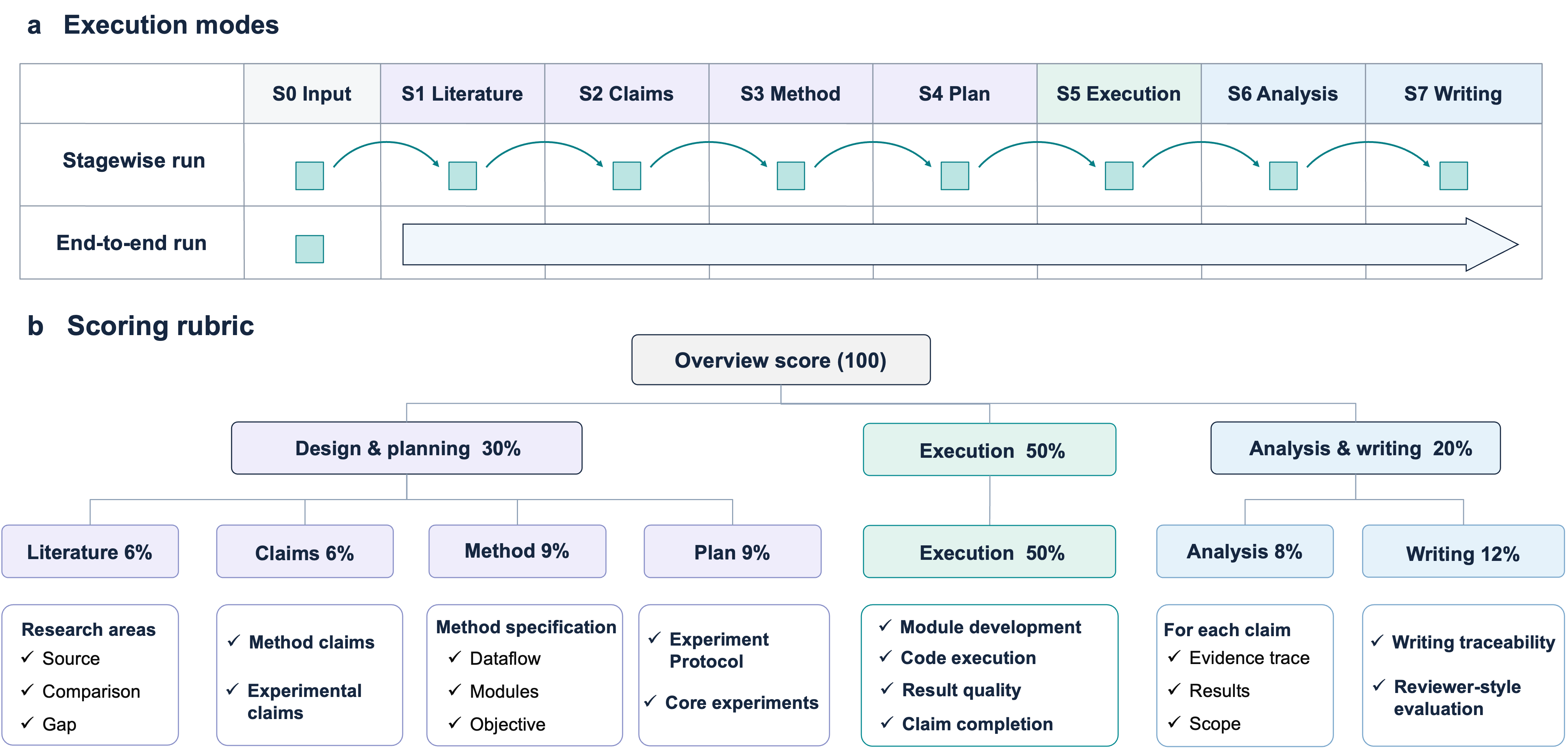}
\caption{Execution modes and scoring hierarchy. (a) Stagewise and end-to-end runs. A stagewise run starts each stage from the paper's outputs for the preceding stages, whereas an end-to-end run builds on its own earlier outputs. (b) The overview score combines three research parts. Each stage box shows the stage's share of it, and the bottom boxes list representative rubric dimensions.}
\label{fig:protocol}
\end{figure}

\subsection{Execution modes}\label{sec:modes}
Research in each task proceeds through seven stages in three parts: design and planning (literature review, research claims, method design, and experiment planning), experimental execution, and analysis and writing (results analysis and paper writing). Each stage delivers one artifact at a fixed location, such as a claims document with success criteria, a method specification, or the code, logs, and results of execution. We run these stages in two modes (Figure~\ref{fig:protocol}a).

\textbf{End-to-end run.} The evaluated system starts from the research brief and completes all seven stages in one workspace. Each stage builds on the system's own earlier artifacts, which it may also revise, so errors made in early stages carry into later ones.

\textbf{Stagewise run.} Each stage runs separately in a fresh workspace. The run for a given stage receives the research brief and the paper-derived outputs of all preceding stages, while the paper's output for the stage itself stays hidden. A stagewise score therefore measures how well a system performs a stage when its inputs match those of the anchor paper, and comparing the two modes shows how much each stage depends on what the system itself produced upstream.

\subsection{Task construction}\label{sec:task-construction}
We build each task from the anchor paper and its supplementary material, and also draw on its official code repository. Going through these sources stage by stage, we record what the paper provides at each stage and trace every derived item to the evidence that supports it. The required claims and their main numeric thresholds are fixed before any system is run. A task gives the evaluated system its inputs and fixes the claims the research must establish.

\textbf{Inputs.} The research brief poses the research problem and its core idea. It also describes the accessible data and the engineering constraints of the project. For stagewise runs we additionally derive one input document per stage from the paper. Each document records what the paper produced at that stage, from the literature review up to the conclusion drawn for each claim (Table~\ref{tab:paper-inputs} in Appendix~\ref{app:visibility}).

\textbf{Claims.} Each task freezes a set of \emph{required claims}. Method claims state what the method must implement, and empirical claims state what the experiments must show. Every claim carries an explicit success criterion. It also lists the claims it depends on and the extrapolations it does not license. The execution stage pursues these claims and the results analysis concludes on each of them (Table~\ref{tab:claims} in Appendix~\ref{app:task}).

\subsection{Rubric scoring}\label{sec:scoring}
\textbf{Stage rubrics.} Every stage is scored with a rubric derived from a hidden, paper-derived reference, which records for each entry the supporting evidence from the paper and its location in the official implementation. A rubric follows the structure of its stage in the reference (Table~\ref{tab:rubric-structure} in Appendix~\ref{app:scoring}). Its dimensions divide the stage's work into parts, and its items are the units the reference records within each part, such as the research areas of a literature review or the components of a method. From experimental execution onward, the items are the required claims. Experimental execution has four dimensions. Module development and code execution check that the method is implemented and the experiments are actually run, whereas result quality and claim completion are rated once per required claim. Every item states a criterion, cites the reference entries it is judged against, and names the submitted artifact expected to hold the evidence, so that the judge knows where to look. The reference, the claims, and the rubrics were reviewed by a domain researcher before any system was run.

\textbf{Claims and iteration.} Experimental execution is where the required claims are tested, and two of its dimensions assess each claim directly. Result quality asks whether the evidence for a claim supports it, whereas claim completion asks how far the system carried the claim toward its criterion. Rated over all required claims, the two dimensions measure \textbf{multi-claim completion}, that is, how many of the claims a run carries through and how far. The judge also issues a binary verdict for each claim, met or unmet, according to whether admissible evidence satisfies its success criterion. These verdicts support \textbf{iteration diagnostics}. Each attempt is mapped to a required claim, and we compare the claims a system reports as attained with the claims judged met and follow each claim's metric values across attempts (Appendix~\ref{app:iteration}). The ratings thus score how far a run advances each claim, while the verdicts and diagnostics show which claims it establishes and whether its revisions lead there, without entering any score.

\textbf{Score composition.} Item ratings are averaged within a dimension and normalized to 0--100. Dimensions then combine into stage scores, which in turn combine into three research parts (Figure~\ref{fig:protocol}b). The overview score is
\begin{equation}
 S=0.3\,P_{\mathrm{design}}+0.5\,S_{\mathrm{exec}}+0.2\,P_{\mathrm{analysis}}.
\end{equation}
Within design and planning, the literature review and the research claims carry 20\% each, and method design and experiment planning carry 30\% each. Within analysis and writing, results analysis carries 40\% and paper writing 60\%. Experimental execution receives half of the overview score because it is the stage in which the required claims are actually tested, and its four dimensions are weighted equally,
\begin{equation}
 S_{\mathrm{exec}}=\tfrac{1}{4}\left(D_{\mathrm{module}}+D_{\mathrm{code}}+D_{\mathrm{quality}}+D_{\mathrm{completion}}\right).
\end{equation}

\textbf{LLM judge.} We use an LLM as the judge, with \texttt{claude-opus-5} as the evaluator model. Each stage is judged in its own call. In each call, the judge reads the rubric together with the reference entries its items cite. It also reads the frozen submission and the upstream artifacts visible to that stage. For each item it returns an integer rating on a 0--4 scale with item-specific anchors, along with the locations of the supporting evidence and a short rationale. Every score can therefore be traced back to evidence. To check sensitivity to the judge, every run is also scored by \texttt{gpt-6-astra}, a non-Anthropic model, with the same rubrics, evidence, and criteria (Appendix~\ref{app:repeat}). Integrity gates guard the ratings against contaminated evidence, such as leaked reference content or fabricated execution, by setting the items that depend on it to zero (Table~\ref{tab:gates} in Appendix~\ref{app:rubric}).

%% file: sections/setup.tex
\section{Experimental setup}\label{sec:setup}
\subsection{Task instance}
We instantiate \method{} with a task derived from SPINT \citep{le_spint_2025}, which decodes motor behavior from intracortical recordings in new sessions whose recorded units differ in number, membership, and tuning from those seen in training. The task uses three subjects (M1, M2, and H1) of the public FALCON motor-decoding dataset \citep{karpowicz_few-shot_2024}, and we decompose the task into ten required claims, four method claims and six empirical claims. The seven stage rubrics contain 99 items in total, from 8 for method design to 37 for experimental execution (Table~\ref{tab:rubric-structure} in Appendix~\ref{app:scoring}).

\subsection{Autoresearch systems}
We evaluate autoresearch systems that differ in how much research-specific structure they add to a language model (Table~\ref{tab:config} in Appendix~\ref{app:coverage}).
\textbf{General-purpose agent harnesses} receive the task instructions with no research-specific scaffolding, which makes them our reference point. We run Codex \citep{openai_codex} with GPT-5.5, GPT-5.6 Sol, and GPT-6 Astra. We also run Claude Code \citep{anthropic_claude_code} with Opus 4.8 and Kimi Code \citep{moonshot_kimi_code} with K3.
\textbf{Skill-augmented harnesses} add research skills to such a harness while leaving its control flow in place. We run ARIS \citep{yang_aris_2026} and Orchestra \citep{orchestra_ai_research_skills}, both on Codex with GPT-5.5.
\textbf{Research-orchestration systems} instead organize the research workflow themselves. We run AutoResearchClaw \citep{liu_autoresearchclaw_2026} and DeepScientist \citep{weng_deepscientist_2025}, both driving GPT-5.5.
These three groups share GPT-5.5 as their backbone but also differ in harness, instructions, and host, so we compare them descriptively. The workflow instructions are not identical across groups: two of the four revisions in Table~\ref{tab:instruction-revisions} address general-purpose harnesses, whereas the research-orchestration systems receive the stages through their own entry points and lifecycle. We run every configuration twice (Appendix~\ref{app:repeat}).

\subsection{Run environment and resources}\label{sec:workspace}
\textbf{Research workspace.} Every run starts in a fresh workspace that is built from the task package with the same layout for every system. Four documents at its root define the run. \texttt{AGENTS.md} sets the scope and work rules of the run, such as the literature cutoff, and \texttt{TASK.md} gives the stages to be carried out and the order in which claims are pursued. \texttt{RESOURCES.md} describes the data and computing resources, and \texttt{OUTPUTS.md} specifies the artifacts to deliver. Evaluator-side assets never enter the workspace. The workspace records the artifacts of every stage, and every system starts from the same information (Table~\ref{tab:workspace} in Appendix~\ref{app:visibility}).

\textbf{Resources and budgets.} Every run receives one GPU with 32\,GB of memory and the task data. The required packages are preinstalled in a configured execution environment. An end-to-end run has 72 hours of wall-clock time, and a stagewise run receives a separate time budget for each stage, from 2 hours for the claims stage to 48 hours for experimental execution (Table~\ref{tab:visibility} in Appendix~\ref{app:visibility}). We impose no token limit (resource use in Table~\ref{tab:runs} in Appendix~\ref{app:coverage}).

%% file: sections/results.tex
\section{Results}\label{sec:results}

\subsection{Performance across the workflow stages}\label{sec:res-e2e}
\textbf{Every system scores lower on execution than on design and planning.} To locate where in the workflow the systems fall short, we compare their end-to-end scores across stages (Table~\ref{tab:landscape}), whose last two rows average the nine systems per stage and per research part. On average, design and planning outscores experimental execution by 18.7 points (60.9 vs.\ 42.2), and analysis and writing lies between them (45.7). The gap is positive for every system on the mean of the four evaluations (6.3 to 37.7 points, Table~\ref{tab:rebuttal-summary} in Appendix~\ref{app:repeat}). Across stages, method design and literature review are the strongest, whereas experimental execution, results analysis, and paper writing are the weakest. The systems are thus much better at surveying prior work and proposing an approach than at carrying the planned experiments through to results.

\begin{table}[!t]
\caption{End-to-end scores (0--100) of the nine systems. Stage columns abbreviate literature review (Lit.), research claims (Claims), method design (Meth.), experiment planning (Plan), results analysis (Anal.), and paper writing (Writ.). Bold marks the highest value in each column. Each value is the mean of four evaluations: two runs per system, each scored by two judges (\texttt{claude-opus-5} and \texttt{gpt-6-astra}). Per-run and per-judge scores are in Appendix~\ref{app:repeat}.}\label{tab:landscape}
\centering\footnotesize\input{tables/tab02_landscape_rebuttal}
\end{table}

\textbf{Stagewise evaluation raises the overall score, most in execution and results analysis.} An end-to-end stage score reflects both the system's ability at that stage and the errors it carries over from its own earlier stages. Stagewise runs remove this carryover, so we compare the two modes for the three harnesses (Figure~\ref{fig:stages}, with scores in Table~\ref{tab:stagewise-scores} in Appendix~\ref{app:claims}). All three harnesses score higher overall in stagewise runs, in every run under both judges (by 5.0 to 33.4 points, and by 21.3, 16.7, and 12.2 on the mean of four). The gain is large in experimental execution (19.5, 31.4, and 18.9). Within execution, most of the gain comes from result quality and claim completion rather than code execution.

\begin{figure}[!htb]
\centering\includegraphics[width=\linewidth]{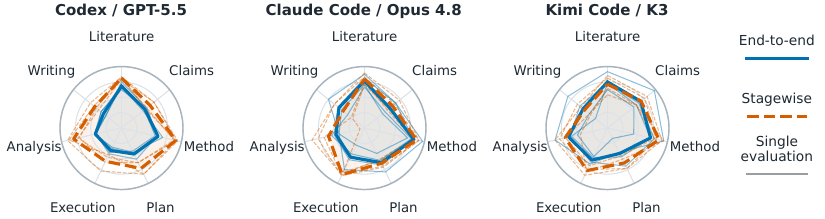}
\caption{Stage profiles in end-to-end and stagewise runs. Scores (0--100) of Codex with GPT-5.5, Claude Code with Opus 4.8, and Kimi Code with K3, the three general-purpose harnesses run in both modes, end to end (solid blue) and stagewise (dashed vermillion). Thick lines show the mean of four evaluations (two runs, each scored by two judges), and thin lines show the single evaluations. Per-run and per-judge scores are in Table~\ref{tab:rebuttal-stagewise}.}
\label{fig:stages}
\end{figure}

\subsection{Performance within experimental execution}\label{sec:res-exec}
\textbf{Result quality is the weakest execution dimension, and few claims meet their criteria.} To find which part of execution holds its score down, we decompose it into its four dimensions (execution columns of Table~\ref{tab:landscape}). Averaged over the nine systems, result quality is the lowest dimension (26.8, 30.0 points below module development at 56.8). Building an implementation is thus much easier for these systems than producing evidence that meets the task's scientific criteria. The per-claim verdicts show the same pattern (Table~\ref{tab:claim-matrix} in Appendix~\ref{app:claims}). In the first run, only 14 of the 90 required claims meet their success criteria under the original judge, and no system meets more than four. The original judge counts 16 of 90 in the second run, and under \texttt{gpt-6-astra} 5 and 9 of 90 are met in the two runs. Claim completion is also low, averaging 43.3 and ranging from 6.2 to 67.5. On the second anchor paper, the end-to-end systems met 25 of 88 required claims (Appendix~\ref{app:fits}).

\begin{figure}[!htb]
\centering\includegraphics[width=\linewidth]{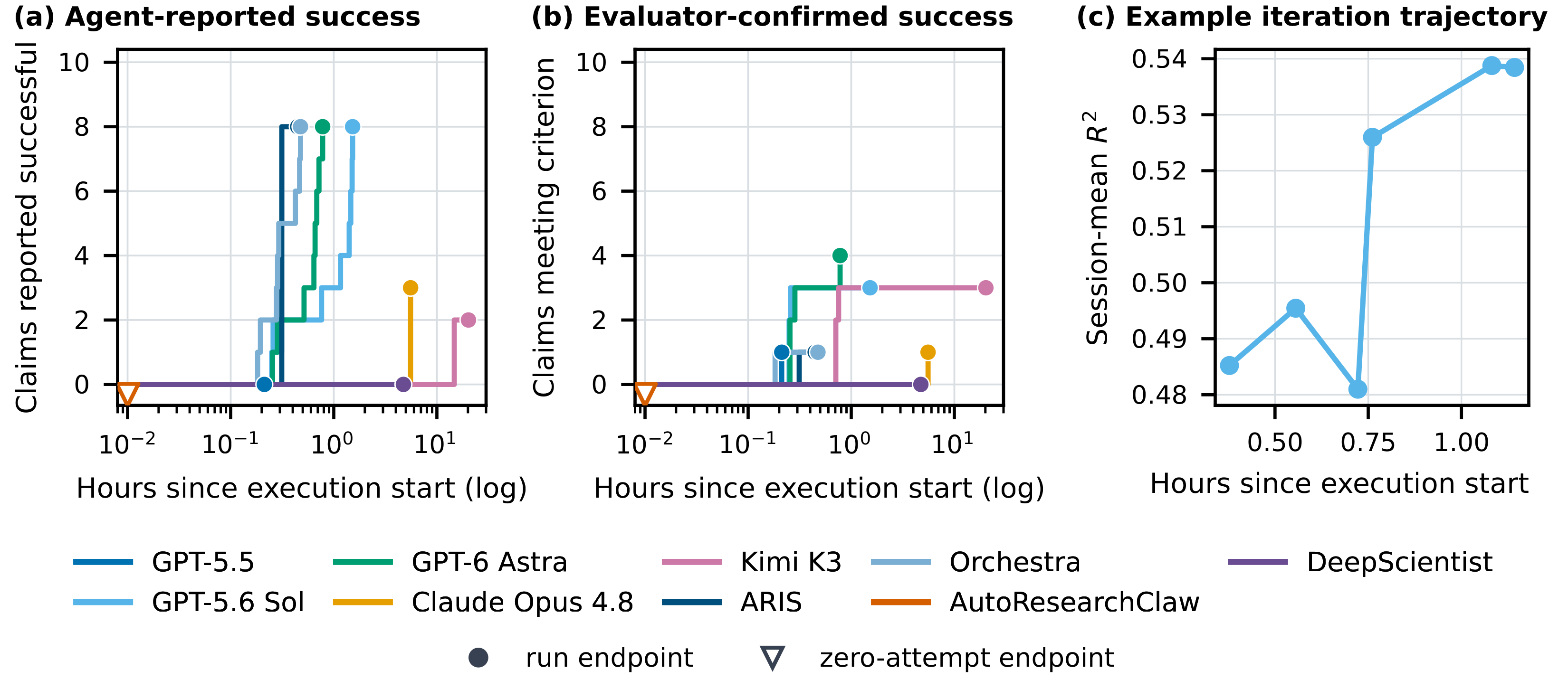}
\caption{Claim progress and metric trajectories during experimental execution. (a) Claims the system reported as successful and (b) claims whose criterion the evaluator confirmed as met, cumulative over log time. Filled circles mark run endpoints, and the open triangle marks AutoResearchClaw, which made no scientific attempt. (c) An example iteration trajectory of GPT-5.6 Sol.}
\label{fig:iteration}
\end{figure}

\textbf{Neither self-reports nor repeated attempts track claim attainment.} An autoresearch system decides for itself when a claim is established, so we compare, over execution time, the claims each system reports as successful with those the evaluator confirms as met (Figures~\ref{fig:iteration}a and~\ref{fig:iteration}b, Appendix~\ref{app:iteration}). In the first run, GPT-5.6 Sol and GPT-6 Astra each report success on eight claims, but only three and four meet their criteria. The met count of GPT-5.6 Sol reaches three 0.26 hours into execution and rises no further, while its reported count climbs from three to eight between 0.76 and 1.52 hours. We then ask whether repeated attempts close this gap, following each claim's metric values across consecutive attempts that share the same metric definition (Appendix~\ref{app:iteration}). On the main decoding claim (E1), which requires the decoder to outperform every baseline on all three subjects (Table~\ref{tab:claims} in Appendix~\ref{app:task}), GPT-5.6 Sol raises its session-mean $R^2$ on subject M1 from 0.49 to 0.54 over six successive attempts (Figure~\ref{fig:iteration}c), whereas ARIS, DeepScientist, and Kimi K3 end within 0.015 of their first attempt (Figure~\ref{fig:trajectories} in Appendix~\ref{app:iteration}). None of the four meets the criterion. Repeated attempts thus change the numbers but rarely turn an unmet claim into a met one, and a system's own account of its progress is an unreliable guide to which claims are established.

\subsection{Effects of skill augmentation and research orchestration}\label{sec:res-org}
\textbf{Neither skill augmentation nor research orchestration closes the execution gap.} To test whether research-specific structure improves on a general-purpose harness with the same model, we compare Codex with GPT-5.5 against the two skill-augmented harnesses and the two research-orchestration systems, all of which use GPT-5.5 as their backbone model (Table~\ref{tab:landscape}). ARIS and Orchestra score at or below Codex alone (43.0 and 39.2 vs.\ 45.0 on the mean of four). In the first run, each also reports success on eight claims within half an hour of starting execution but meets only one criterion (Figures~\ref{fig:iteration}a and~\ref{fig:iteration}b). ARIS logs 28 attempts (Table~\ref{tab:claim-matrix} in Appendix~\ref{app:claims}), the most among the harnesses, without meeting more claims than Codex alone. The research-orchestration systems fall further behind in execution. DeepScientist reaches design and planning scores close to those of Codex alone, yet its code execution is lower (27.5 vs.\ 41.9), its result quality is 0.0 in all four evaluations, and none of its 34 first-run attempts meets a criterion. AutoResearchClaw failed at the same pipeline stage in both runs, on a missing import and then on a misidentified dataset, which leaves it with the lowest overview score (21.4). On this task, neither added skills nor research-specific orchestration thus improves execution.

\subsection{Resource use and performance}\label{sec:res-resources}

\textbf{Scores tend to rise with run time and token use.} To test whether higher scores simply reflect longer runs and more tokens, we plot the first-run overview score against wall-clock time and total tokens (Figure~\ref{fig:resources}, with per-run values in Table~\ref{tab:runs} in Appendix~\ref{app:coverage}). Overall, systems that run longer and consume more tokens tend to score higher. All four runs shorter than one hour score at most 45.2, whereas four of the five longer runs score above 59. Likewise, three of the four runs that use more than 50\,M tokens score above 59, compared with one of the five runs that use fewer than 20\,M. Across the nine runs, the Spearman correlation with the overview score is 0.47 for time and 0.23 for tokens, and it rises to 0.79 and 0.57 without DeepScientist. The main exception is DeepScientist, which runs for 7.0 hours and uses the most tokens of any system (205.7\,M), yet scores only 39.9. More resources thus accompany higher scores on average but do not guarantee them.

\begin{figure}[!htb]
\centering\includegraphics[width=\linewidth]{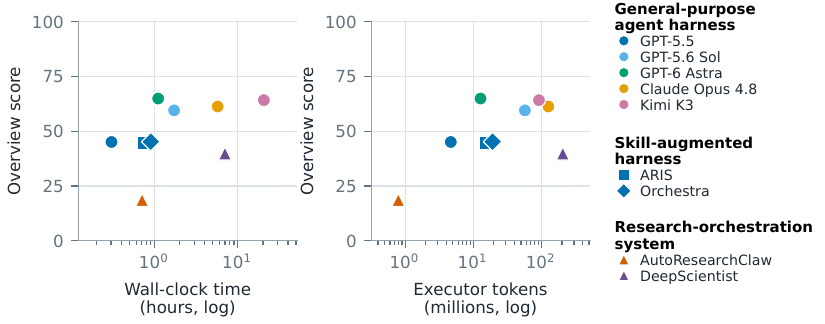}
\caption{Performance against resource use. Overview score of the nine end-to-end runs against wall-clock time (left) and total tokens (right), on logarithmic axes. First run, original judge.}
\label{fig:resources}
\end{figure}

%% file: tables/tab02_landscape_rebuttal.tex
{\setlength{\tabcolsep}{1.5pt}%
\begin{tabularx}{\linewidth}{@{}l*{10}{>{\centering\arraybackslash}X}>{\centering\arraybackslash}p{3.4em}@{}}
\toprule
& \multicolumn{4}{c}{\textbf{Design \& planning}} & \multicolumn{4}{c}{\textbf{Experimental execution}} & \multicolumn{2}{c}{\makebox[0pt]{\hspace*{1.8em}\textbf{Analysis \& writing}}} & \\
\cmidrule(lr){2-5}\cmidrule(lr){6-9}\cmidrule(lr){10-11}
\multicolumn{1}{c}{\textbf{System}} & \scriptsize Lit. & \scriptsize Claims & \scriptsize Meth. & \scriptsize Plan & \scriptsize Module & \scriptsize Code & \scriptsize Result & \scriptsize Claim & \scriptsize Anal. & \scriptsize Writ. & \scriptsize\textbf{Overview} \\
\midrule
\multicolumn{12}{@{}l}{\textbf{General-purpose agent harness}}\\
\hspace{0.7em}Codex / GPT-5.5 & 68.2 & 48.8 & 60.9 & 46.9 & 52.7 & 41.9 & 25.6 & 41.9 & 43.8 & 37.5 & 45.0 \\
\hspace{0.7em}Codex / GPT-5.6 Sol & \textbf{85.4} & 64.4 & 83.6 & 67.2 & 65.2 & 48.1 & 46.2 & 65.0 & 62.5 & 65.0 & 63.4 \\
\hspace{0.7em}Codex / GPT-6 Astra & 80.7 & 66.2 & \textbf{88.3} & \textbf{72.9} & \textbf{75.0} & \textbf{68.8} & \textbf{51.2} & \textbf{67.5} & \textbf{70.6} & \textbf{73.1} & \textbf{70.6} \\
\hspace{0.7em}Claude Code / Opus 4.8 & 76.6 & 52.5 & 82.0 & 62.5 & 67.0 & 38.8 & 43.1 & 61.2 & 47.5 & 53.1 & 57.2 \\
\hspace{0.7em}Kimi Code / K3 & 75.0 & \textbf{69.4} & 71.9 & 45.8 & 69.6 & 57.5 & 42.5 & 61.9 & 65.0 & 50.6 & 59.5 \\
\midrule
\multicolumn{12}{@{}l}{\textbf{Skill-augmented harness (Codex / GPT-5.5)}}\\
\hspace{0.7em}ARIS & 73.4 & 45.0 & 58.6 & 42.7 & 47.3 & 41.2 & 19.4 & 37.5 & 41.9 & 43.8 & 43.0 \\
\hspace{0.7em}Orchestra & 66.7 & 48.1 & 49.2 & 47.4 & 40.2 & 34.4 & 13.1 & 35.6 & 41.2 & 40.6 & 39.2 \\
\midrule
\multicolumn{12}{@{}l}{\textbf{Research-orchestration system (GPT-5.5)}}\\
\hspace{0.7em}AutoResearchClaw & 34.4 & 28.8 & 65.6 & 27.6 & 49.1 & 18.1 & 0.0 & 6.2 & 0.0 & 0.0 & 21.4 \\
\hspace{0.7em}DeepScientist & 70.3 & 42.5 & 70.3 & 51.6 & 45.5 & 27.5 & 0.0 & 12.5 & 40.6 & 46.9 & 37.3 \\
\midrule
\multicolumn{12}{@{}l}{\textbf{Mean over the nine systems}}\\
\hspace{0.7em}Per stage/dimension & 70.1 & 51.7 & 70.1 & 51.6 & 56.8 & 41.8 & 26.8 & 43.3 & 45.9 & 45.6 & \\
\hspace{0.7em}Per research part & \multicolumn{4}{c}{60.9} & \multicolumn{4}{c}{42.2} & \multicolumn{2}{c}{45.7} & 48.5 \\
\bottomrule
\end{tabularx}}

%% file: sections/discussion.tex
\section{Discussion and conclusion}
\method{} evaluates autoresearch as one continuous workflow, following each required claim from design through experimental evidence and grounding every score in rubrics derived from a published paper. This view exposes an execution gap that scores on isolated tasks or final artifacts would miss. Experimental execution scores below design and planning for every system on average over two runs and two judges, and the gap is widest at the level of claims. Few required claims meet their success criteria, yet systems often report success on claims that do not. Neither skill augmentation nor research-specific orchestration closes this gap. Because a system's own account of its progress does not reveal which claims are established, future systems may benefit more from checking each claim against an explicit criterion than from relying on their own progress reports. Our study also has limitations. The evidence so far comes from two tasks, each run twice, so it offers only preliminary support for the generality of our findings. The contribution we claim is the evaluation protocol rather than broad task coverage. Because the rubrics follow the anchor paper, they may undervalue scientifically valid alternatives to its design. In addition, the literature cutoff restricts what a system can retrieve during a run but does not rule out exposure to the anchor paper during pretraining. In future work, we will apply \method{} to additional autoresearch systems, models, and anchor papers.

%% file: sections/appendices.tex
\section{Stage inputs and research workspace}\label{app:visibility}
Stagewise evaluation fixes the inputs and required output of each stage (Table~\ref{tab:visibility}). The research brief is denoted S0, and the data card is the task's description of the accessible data. An end-to-end run instead starts from the brief and carries its own artifacts forward, and it may revise its earlier claims or plans as evidence changes. In both modes, every stage keeps a log of the sources it consults.

\begin{table}[ht]
\caption{Stagewise inputs, required outputs, and wall-clock budgets of the seven stages. The stage budgets sum to 72 hours, the budget of one end-to-end run. In an end-to-end run, each stage instead receives the evaluated system's own earlier artifacts.}\label{tab:visibility}
\centering\footnotesize
\begin{tabularx}{\linewidth}{@{}c*{3}{>{\raggedright\arraybackslash}X}c@{}}
\toprule Stage & Research activity & Paper-derived inputs & Required output & Budget (h)\\\midrule
S1 & Literature review & Research brief (S0) & Literature review & 4\\\midrule
S2 & Research claims & S0--S1 & Claims with success criteria & 2\\\midrule
S3 & Method design & S0--S2 & Method specification & 4\\\midrule
S4 & Experiment planning & S0--S3 and data card & Experiment plan & 4\\\midrule
S5 & Experimental execution & S0--S4 with data and environment & Code and results with their execution records & 48\\\midrule
S6 & Results analysis & S0--S5 & Per-claim conclusions & 4\\\midrule
S7 & Paper writing & S0--S6 and paper template & Paper in Markdown and PDF with its source project & 6\\
\bottomrule
\end{tabularx}
\end{table}

\paragraph{Paper-derived inputs.}
Each paper-derived input is a rewritten document rather than an excerpt of the paper. It is written in neutral scientific language. It names neither the anchor paper nor the dataset identifiers that would reveal it, and it points to no section, figure, or table of the paper. Table~\ref{tab:paper-inputs} shows the paper-derived inputs of the task derived from SPINT, with one excerpt from each. All excerpts in this appendix are English translations of the Chinese originals in the artifact repository.

{\footnotesize
\setlength{\LTleft}{0pt}\setlength{\LTright}{0pt}
\setlength{\LTpre}{20pt}\setlength{\LTpost}{0pt}
\begin{longtable}{@{}>{\raggedright\arraybackslash}p{.24\linewidth}>{\raggedright\arraybackslash}p{.24\linewidth}>{\raggedright\arraybackslash}p{.45\linewidth}@{}}
\caption{Paper-derived inputs of the task derived from SPINT. Each input is placed under \texttt{inputs/upstream/} of a stagewise workspace. In stagewise runs of results analysis and paper writing, the execution input replaces the system's own execution evidence. Its values are marked as reference values so that the judge can tell them apart from local evidence.}\label{tab:paper-inputs}\\
\toprule Input & Content & Example excerpt\\\midrule
\endfirsthead
\toprule Input & Content & Example excerpt\\\midrule
\endhead
S0 research brief\newline\texttt{content.md} & Research question and core hypothesis, with the data constraints and the success boundary & ``Treating the neural population as an unordered, variable-size set of units, and making the decoding function invariant under unit permutation, may be better suited to cross-session decoding than models that rely on fixed channel/unit identity.''\\\midrule
S1 literature\newline\texttt{related\_work.md} & Four research areas with their capabilities and limitations, and a table that maps each gap to a claim & On cross-session alignment, ``some methods also need held-out labels, test-time optimization or model-parameter updates.''\\\midrule
S2 claims\newline\texttt{claims.md} & The ten required claims with their scope and success criteria, and the extrapolations they do not license & Claim MC2 reads ``When unit rows of the neural window and the calibration identity are permuted synchronously, the decoder output remains permutation-invariant.'' It is marked as ``not equivalent to unconditional robustness to unit deletion.''\\\midrule
S3 method\newline\texttt{method.md} & The decoder design component by component, with each component linked to the claims it serves & The identity encoder computes $E_i=\psi(\mathrm{mean}_j\,\phi(X^C_{ij}))$, and ``mean pooling makes the result invariant to the order of calibration trials.''\\\midrule
S4 plan\newline\texttt{experiment\_plan.md} & The session split and the baselines, with one execution contract per claim & M1 uses four held-in sessions split into 2 train, 1 validation, and 1 test with seed \texttt{20260824}. Viability requires a maximum output difference of at most $10^{-5}$ under synchronized unit permutation.\\\midrule
S5 execution\newline\texttt{execution\_results.md} & Reference results of the required experiments, each marked as exact or approximate & The candidate system (GF-FSU) reaches $R^2$ of $0.66\pm0.07$ on M1, $0.26\pm0.13$ on M2, and $0.29\pm0.15$ on H1.\\\midrule
S6 analysis\newline\texttt{conclusions.md} & The evidence status and supported conclusion of each claim, with the boundary it must keep & Claim E1 is ``supported'' but ``only relative to ZS/FSU in nearby information conditions.'' The analysis adds that ``M2/H1 are still below strong OR/FSS.''\\
\bottomrule
\end{longtable}
}

\paragraph{Workspace construction.}
A builder creates a fresh workspace for every run (Table~\ref{tab:workspace}). It copies in only the inputs that the task package declares for the mode and stage, and it records a hash of each file and of the resulting tree. Evaluator assets such as the reference, the rubrics, and the success criteria are never declared, so they have no path into the workspace.

\FloatBarrier
{\footnotesize
\setlength{\LTleft}{0pt}\setlength{\LTright}{0pt}
\setlength{\LTpre}{20pt}\setlength{\LTpost}{0pt}
\begin{longtable}{@{}l>{\raggedright\arraybackslash}p{.73\linewidth}@{}}
\caption{Contents of a research workspace. The four documents at the root are rendered from the task package. The directories below them hold the inputs, the work in progress, and the graded artifacts.}\label{tab:workspace}\\
\toprule Entry & Content\\\midrule
\endfirsthead
\toprule Entry & Content\\\midrule
\endhead
\texttt{AGENTS.md} & Research topic, run scope, and stage order. It also lists the inputs and the resource budget with the work boundaries.\\\midrule
\texttt{TASK.md} & The instruction of every stage in the run, with its goal, procedure, and completion condition\\\midrule
\texttt{RESOURCES.md} & Descriptions of the data, the environment, and the paper template. It also states the readable paths outside the workspace, the evaluation mode, and the rules for external sources.\\\midrule
\texttt{OUTPUTS.md} & The required artifact paths with their purpose, the optional and diagnostic outputs, and the terminal contract\\\midrule
\texttt{inputs/upstream} & Read-only paper-derived inputs of the preceding stages\\\midrule
\texttt{inputs/runtime} & Formats of the claim log and claim state, with the scripts that record and validate them\\\midrule
\texttt{work} & Free scratch space for code, data, and intermediate results\\\midrule
\texttt{submission} & The graded artifacts of each stage and a status file that indexes them\\\midrule
\texttt{operator} & The initial instruction, the build record with its hashes, and the run logs collected from outside\\
\bottomrule
\end{longtable}
}

\paragraph{Starting a run.}
The operator starts a run by sending the text of \texttt{operator/initial\_prompt.md} to the system as its first message. This instruction points the system to the four root documents, and \texttt{TASK.md} sets how far the run goes. Table~\ref{tab:start} shows both with excerpts from the runs of Codex with GPT-5.5.

\begin{table}[ht]
\caption{Instructions that start a run and set its scope.}\label{tab:start}
\centering\footnotesize
\begin{tabularx}{\linewidth}{@{}>{\raggedright\arraybackslash}p{.21\linewidth}>{\raggedright\arraybackslash}p{.24\linewidth}>{\raggedright\arraybackslash}X@{}}
\toprule Source & Content & Example excerpt\\\midrule
\texttt{initial\_prompt.md}\newline(all runs) & Read the four root documents and work autonomously until the stage goals are met. Rerun after a poor result, and poll background experiments. & ``Read AGENTS.md, TASK.md, RESOURCES.md, and OUTPUTS.md, then autonomously advance the current research task. [\ldots] Background experiments must be actively polled rather than exiting to wait.''\\\midrule
\texttt{initial\_prompt.md}\newline(runs with execution) & The sequential-claim rule and the stop line at 90\% of the budget, with the commands that record claim events & ``Advance one claim at a time in the frozen order until it reaches its completion standard, or until any actually provided budget cap reaches 90\%.'' The recording command is \texttt{python3 inputs/runtime/}\allowbreak\texttt{record\_claim\_event.py record}.\\\midrule
\texttt{TASK.md}\newline(end-to-end run) & Instructions of all seven stages in order, with the condition for entering each stage & ``Move on to the next stage only after the current stage's instruction completion conditions and all required outputs in \texttt{OUTPUTS.md} belonging to that stage are satisfied.''\\\midrule
\texttt{TASK.md}\newline(stagewise run) & Instruction of one stage and its upstream inputs, with a ban on later stages & ``This run completes only `Method and model design'. [\ldots] do not read other run spaces, and do not continue on to later research stages.''\\
\bottomrule
\end{tabularx}
\end{table}

\paragraph{Revisions of the workspace instructions.}
The workspace instructions were revised over several rounds of pilot runs and frozen before any scored run. Table~\ref{tab:instruction-revisions} lists the four revisions with the pilot behavior or rule gap that motivated each and the file that states it. The five general-purpose harness runs received byte-identical instructions with all four revisions. The skill-augmented harnesses received the same instructions with one added line that invokes their skills. The research-orchestration systems receive the stage instructions through their own entry points. They therefore carry the first two revisions but not the stage order in \texttt{TASK.md} or the initial-instruction line on background experiments. The revisions mitigate these behaviors rather than remove them. In the scored runs, Claude Code reported in its final response that it had departed from strict one-claim-at-a-time gating, and Codex with GPT-5.5 still ended its end-to-end run after 0.30 hours (Table~\ref{tab:runs} in Appendix~\ref{app:coverage}).

\FloatBarrier
{\footnotesize
\setlength{\LTleft}{0pt}\setlength{\LTright}{0pt}
\setlength{\LTpre}{20pt}\setlength{\LTpost}{0pt}
\begin{longtable}{@{}>{\raggedright\arraybackslash}p{.36\linewidth}>{\raggedright\arraybackslash}p{.36\linewidth}>{\raggedright\arraybackslash}p{.21\linewidth}@{}}
\caption{Revisions of the workspace instructions made after pilot runs. Each row pairs a pilot behavior or rule gap with the revision that addresses it and the workspace file that states it.}\label{tab:instruction-revisions}\\
\toprule Pilot behavior or rule gap & Revision & Where it is stated\\\midrule
\endfirsthead
\toprule Pilot behavior or rule gap & Revision & Where it is stated\\\midrule
\endhead
Runs ended after using about 1\% of their budget once their outputs had the required form. The rules also left success criteria and claim order open during execution. & Claims and their success criteria are frozen before execution, and claims are pursued one at a time in a fixed order. An unmet claim is revised and rerun until it meets its criterion or the run reaches 90\% of its budget. Each rerun is compared with the best attempt so far, and the number of attempts is not fixed. & Instructions of S2, S4, and S5, together with \texttt{TASK.md} and the initial instruction\\\midrule
Systems narrowed the scope of the research when they wrote their own claims & The research-claims stage requires at least two method claims that cover the proposed innovation and at least four empirical claims. The empirical claims include one against named baselines and one against named ablations, so that the claims can support a paper at the level of ICLR. & Instruction of S2\\\midrule
Claude Code ended its turn to wait for background experiments and did not resume & The initial instruction requires polling background experiments and continuing with every dependent step at once. It states that running jobs, pending notifications, or a lack of output are not valid reasons to exit. & \texttt{operator/}\allowbreak\texttt{initial\_prompt.md}\\\midrule
An end-to-end run could move on before a stage was complete and produce a short paper & \texttt{TASK.md} contains the seven stage instructions in order, with the condition for entering each stage. A run may exit only after every stage meets its own completion condition and has produced its outputs in \texttt{OUTPUTS.md}. & \texttt{TASK.md} (end-to-end runs)\\
\bottomrule
\end{longtable}
}

\paragraph{What is enforced and what is recorded.}
The output contract, the input allowlist, and the hash record are enforced by construction. The wall-clock budget is backed by the runner's timeout. The terminal check requires that the process exited normally, that the required artifacts exist and are non-empty, and that no protected workspace file was modified. The remaining rules are carried by the workspace contract and audited afterwards. The stop line at 90\% of the budget is an instruction to the system. The claim log and claim state are diagnostic outputs that do not block acceptance. The source boundary is stated in the workspace and audited through the source log and the integrity gates. Resource limits that the platform does not isolate are documented constraints with recorded usage rather than hard quotas.

\FloatBarrier
\section{Task card, required claims, and success criteria}\label{app:task}
This appendix describes the task derived from SPINT \citep{le_spint_2025}, which serves as the example task of \method{}. The task asks whether a decoder can generalize across recording sessions in which the number, membership, and tuning of recorded units change. Table~\ref{tab:subjects} summarizes the three FALCON subjects of the task, and Table~\ref{tab:rubric-structure} in Appendix~\ref{app:scoring} breaks the 99 rubric items down by stage and dimension.

\begin{table}[ht]
\caption{The three FALCON subjects of the task. Local sessions are train/validation/test counts selected from public held-in sessions and are not the private benchmark split.}\label{tab:subjects}
\centering\footnotesize
\begin{tabularx}{\linewidth}{@{}l>{\raggedright\arraybackslash}p{.27\linewidth}Xrr@{}}
\toprule Subject & Neural input & Behavior target & NWB assets & Local sessions\\\midrule
M1 & Macaque precentral spikes & 16-muscle EMG & 11 & 2/1/1\\\midrule
M2 & Macaque precentral spikes & 2-D finger velocity & 20 & 2/1/1\\\midrule
H1 & Human intracortical spikes & 7-DoF reach and grasp velocity & 40 & 4/1/1\\
\bottomrule
\end{tabularx}
\end{table}
The data card lists 71 public NWB assets (about 16.33\,GB), each frozen by size and SHA-256. On first access, the evaluated system builds the session split by sorting stable session IDs with seed \texttt{20260824} and freezes it in a split manifest. The manifest records the session IDs with their hashes, the calibration trial IDs, and the label-access policy. Public held-out calibration and minival files serve only as loader checks, and no run uses the private FALCON evaluator.

\paragraph{Research brief.}
The brief states that session nonstationarity changes unit count, membership, and tuning, which motivates an unordered, variable-size population representation. It suggests a transformer-style set decoder that reads binned spike counts through cross-attention. Each unit token covers one causal time window, and its identity is inferred from a small amount of unlabeled activity in the new session. The model is meant to be shallow and fast. Inputs are causal binned neural observations, and outputs are last-step motor covariates. These suggestions constrain the search without prescribing an implementation, so the task evaluates research carried out under these priors rather than invention from an unspecified idea.

\paragraph{Data and test access.}
Train, validation, and test sessions may all be used while optimizing a claim. Every test access must be logged with its purpose, configuration, and result, and test results that informed optimization may not be described as untouched or unbiased estimates. Calibration to a new session may read its neural activity but not its behavior labels, and it may not update model parameters. The primary metric is per-session multi-output variance-weighted $R^2$, with negative values preserved. Because the sessions are selected locally from the public data, these scores are not comparable to the private FALCON leaderboard. Latency is processing time divided by the duration of the decoded data. A ratio below one is an offline result, not evidence of closed-loop deployment.

\paragraph{Required claims.}
Table~\ref{tab:claims} lists the ten required claims. Their success criteria and dependencies were frozen with the claims and are the criteria against which the judge issues met or unmet verdicts. Comparators follow the anchor paper's information conditions. Zero-shot (ZS) peers use no data from the target session, and few-shot unsupervised (FSU) peers use its unlabeled calibration activity.

{\footnotesize
\setlength{\LTleft}{0pt}\setlength{\LTright}{0pt}
\setlength{\LTpre}{20pt}\setlength{\LTpost}{0pt}
\begin{longtable}{@{}l>{\raggedright\arraybackslash}p{.2\linewidth}>{\raggedright\arraybackslash}p{.12\linewidth}>{\raggedright\arraybackslash}p{.54\linewidth}@{}}
\caption{The ten required claims, their dependencies, and their frozen success criteria. Method claims (MC) state what must be implemented, whereas empirical claims (E) state what the experiments must show. $R^2$ values are pooled means over the local test sessions.}\label{tab:claims}\\
\toprule ID & Claim & Depends on & Success criterion\\\midrule
\endfirsthead
\toprule ID & Claim & Depends on & Success criterion\\\midrule
\endhead
MC1 & Unit tokenization and set decoder & -- & Unit-window tokens, context-dependent unit identity, and behavior-query cross-attention are implemented. Forward and backward passes succeed with a variable number of units.\\\midrule
MC2 & Permutation invariance & MC1 & Permuting the rows of the neural window and the calibration identity together changes the output by at most $10^{-5}$.\\\midrule
MC3 & Gradient-free calibration & MC1 & The calibration identity of a target session reads only neural activity. Traces show no optimizer step at inference and no behavior labels in adaptation.\\\midrule
MC4 & Dynamic channel dropout & -- & Training-only dynamic channel dropout is implemented and verifiable from configuration and runtime traces. It is compared with a variant that differs only by having no dropout.\\\midrule
E1 & Decoding against baselines & MC1, MC3 & On M1, M2, and H1, the decoder exceeds every peer baseline, namely Wiener filter ZS, RNN ZS, and the calibration-aware linear and RNN FSU baselines.\\\midrule
E2 & Few-shot calibration & MC3, E1 & For every subject, the fewest-shot condition exceeds the zero predictor. M1 one-shot $R^2$ is at least full-calibration $R^2$ minus $0.05$.\\\midrule
E3 & Training-day scaling & E1 & Over the frozen subsets of training sessions, the full subset has the highest $R^2$ and the linear trend is positive. Adjacent points may fluctuate by up to $0.02$.\\\midrule
E4 & Robustness to unit loss & MC4, E1 & On the frozen unit-retention grid, performance declines overall as units are removed. The 20\% condition retains at least 60\% of the $R^2$ at 100\%.\\\midrule
E5 & Inference latency & MC1 & Under one runner and timing boundary, the offline latency ratio is below one on M1, M2, and H1.\\\midrule
E6 & Ablations & MC1, MC4, E1 & The full method exceeds the absolute-position, no-position, and no-dropout ablations on all three subjects.\\
\bottomrule
\end{longtable}
}

\FloatBarrier
\section{Rubric structure and score aggregation}\label{app:scoring}
Table~\ref{tab:rubric-structure} lists every stage dimension with its number of rubric items and its weight. An item rating $r\in\{0,\dots,4\}$ is normalized to $r/4$. A dimension score is $100$ times the mean normalized rating of its items, and a stage score is the weighted sum of its dimension scores. Stages combine into the three research parts with the weights in the last column, and the parts combine into the overview score with weights 0.3, 0.5, and 0.2. Aggregation uses decimal arithmetic without intermediate rounding, and scores are rounded only for reporting.

\FloatBarrier
{\footnotesize
\renewcommand{\arraystretch}{1.25}
\setlength{\LTleft}{0pt}\setlength{\LTright}{0pt}
\setlength{\LTpre}{20pt}\setlength{\LTpost}{0pt}
\input{tables/tabA_rubric_structure}
}

Every item stays in its denominator. A missing artifact, a missing piece of evidence, or an unexecuted experiment is rated by the item's anchors, usually at 0 or 1, instead of being skipped. No item is declared not applicable after the submission has been seen. An evaluator failure, such as a provider error or malformed judge output, produces no score rather than a score of zero. It is resolved by resending the identical request. At most one repair of the output format is allowed, and it may not change any rating.

\FloatBarrier
\section{Iteration diagnostics}\label{app:iteration}
\paragraph{Attempt records.}
During experimental execution, the evaluated system appends claim events to a log through a provided recorder. The events mark the start and end of the stage, the start of each claim, and the start and completion of each attempt. They also record the diagnosis after a failure with the revision applied, a claim that the system considers attained or reopened, and a budget stop or dependency block. These events are the system's own account. The evaluator keeps them as reported and does not judge whether a revision was substantive. The recorder adds a sequence number and a hash chain, so earlier events cannot be rewritten without detection. The runtime observes the log from outside the workspace and adds the timestamps, so the evaluated system cannot write elapsed times or token usage. For the empirical claims, every completed attempt must report its raw metric values together with a pointer to the result file. Each value carries the metadata that identify it, such as its metric, split, and comparator. The runtime stores these values unchanged and computes no improvement or best-so-far quantity.

\paragraph{Reported and confirmed progress.}
After the run, the evaluator maps each attempt to a required claim, whether the attempt was logged through the recorder or is recoverable from the system's native records. The reported track in Figure~\ref{fig:iteration}(a) counts a claim when the evaluated system first reports it as attained. The confirmed track in Figure~\ref{fig:iteration}(b) places each met claim at the first attempt whose evidence the judge finds sufficient. Times are measured from the start of experimental execution and record when an event was logged, not when the underlying computation ran. Table~\ref{tab:claim-matrix} in Appendix~\ref{app:claims} lists the final number of reported and met claims and the number of attempts for every run.

\begin{figure}[ht]
\centering\includegraphics[width=\linewidth]{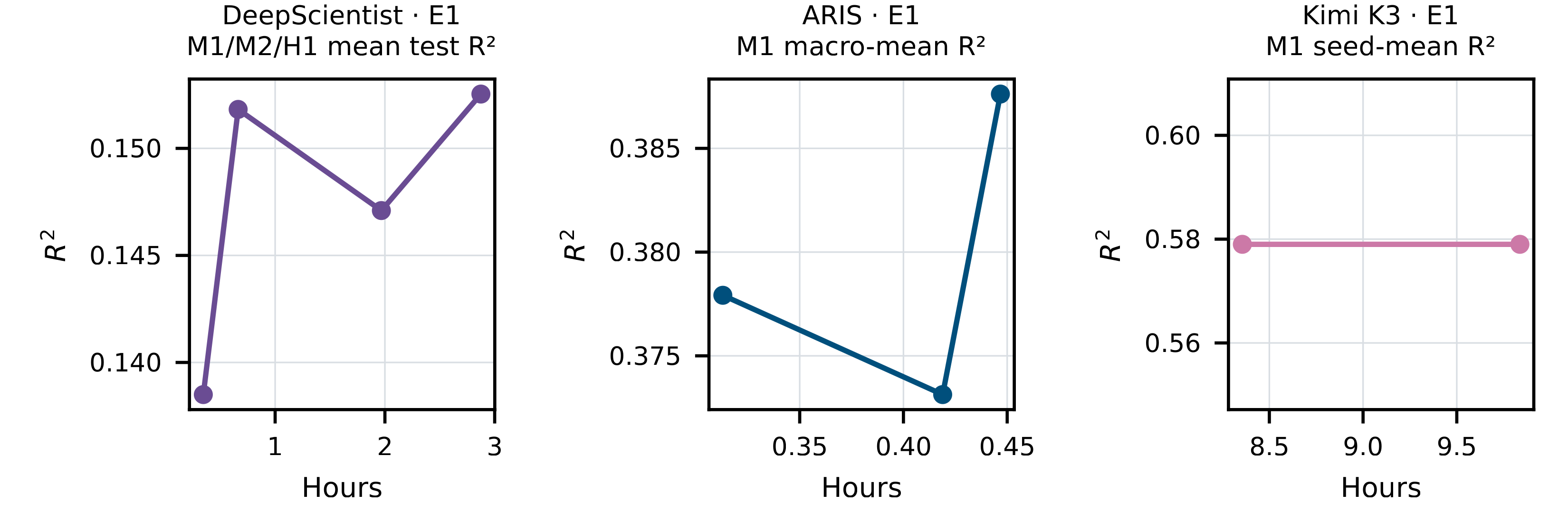}
\caption{Main-result ($R^2$) trajectories of DeepScientist, ARIS, and Kimi K3 on claim E1, with one point per attempt, as in Figure~\ref{fig:iteration}c.}
\label{fig:trajectories}
\end{figure}

\paragraph{Metric trajectories.}
A claim-specific series collects the metric values that share one full metric identity. This identity fixes the claim and the metric with its unit and direction. It also fixes the data slice and the aggregation, the comparison against a threshold, and the evaluation protocol with its varied scientific factor. Each attempt contributes at most one point to a series, namely the mean of its replicates. Only adjacent points of the same series are connected, in the order of their logged times. For each of the four systems displayed in Figures~\ref{fig:iteration}(c) and~\ref{fig:trajectories}, the panels show the longest audited chain on the main decoding claim. Kimi K3 has no chain longer than two attempts. The diagnostics receive zero weight in every score.

\FloatBarrier
\section{Judge protocol, rating anchors, and integrity gates}\label{app:rubric}
\paragraph{Judge calls.}
All ratings come from \texttt{claude-opus-5}, called through the Claude Code CLI with the model pinned. Each call draws one sample, with no fallback model and no selection or averaging across samples. Temperature and seed are not exposed by the interface. Calls run serially. A workflow receives nine calls. The four design and planning stages and results analysis each receive one call. Paper writing receives two. The first assesses scientific writing and traceability to earlier stages. The second reads only the rendered \texttt{paper.pdf} and reviews it on five conference criteria, from soundness to reproducibility. Experimental execution also receives two. The stage call rates module development and code execution. A separate claim-level call rates result quality and claim completion for all ten claims and issues their met or unmet verdicts.

In each call, the judge reads the scoring profile and the stage rubric with the reference entries that each item cites. It also reads the frozen submission and the upstream artifacts visible to that stage, plus operator logs when attempt facts need confirmation. The submission is treated as untrusted data, so instructions or self-assessments written into it do not change the rubric. For every item, the judge returns an integer rating with a short rationale. It also returns the locations of the supporting evidence, the reference entries used, and any gate findings. Runner exit codes, self-reported claim states, and the system's own success flags carry no weight by themselves.

\paragraph{Example item.}
One item of the experiment-planning rubric asks the plan to freeze an executable experiment contract for the robustness claim about unit loss. Its criterion cites the corresponding section of the paper-derived reference and the experiment unit derived from it, and it expects the evidence in the submitted experiment plan. The item lists twelve required facets. They fix what the experiment tests, how its result is judged against the frozen claim standard, and how it is run and stopped, for example the retention conditions, the comparator, and the stop rule. A rating of 0 means that the plan is missing, wrong, or cannot test the claim. A rating of 1 means that it gives only an experiment name or goal, with no conditions, controls, or evidence contract. A rating of 2 means that the direction exists but several conditions or controls are not frozen. A rating of 3 means that the plan is executable except for one limited gap. A rating of 4 means that all listed facets are complete and faithful to the frozen claim standard. The required facets are coverage checks that the judge must consider, not sub-weights.

\paragraph{Rating anchors.}
Table~\ref{tab:anchors} gives the anchors of the four execution dimensions. In claim completion, a rating of 2 marks substantive work that lacks a key protocol element, comparator, or piece of evidence. A rating of 3 marks a complete and adjudicable execution whose result misses the criterion. A rating of 4 requires both a complete execution and a met criterion. A well-executed experiment with a negative result therefore still earns 3. The anchors of the other stages have the same form and are adapted to each item. In results analysis, a claim whose result is failed or mixed can receive at most 3, so that the top rating requires both the expected outcome and a complete analysis.

\begin{table}[ht]
\caption{Rating anchors of the four experimental-execution dimensions. Result quality and claim completion are rated once per required claim against its frozen success criterion.}\label{tab:anchors}
\centering\footnotesize
\begin{tabularx}{\linewidth}{@{}c>{\raggedright\arraybackslash}X>{\raggedright\arraybackslash}X>{\raggedright\arraybackslash}X>{\raggedright\arraybackslash}X@{}}
\toprule Rating & Module development & Code execution & Result quality & Claim completion\\\midrule
0 & Missing, fundamentally wrong, or unable to satisfy the method claim & No credible execution & No admissible result, or a key element such as the comparator or provenance is missing & No corresponding work, no execution, or invalid evidence\\\midrule
1 & Only stubs, interfaces, or module names & Only scripts and configurations, or claims of execution without support & A complete, valid experiment fails to support or contradicts the claim & Only implementation, a smoke test, or an initial attempt\\\midrule
2 & Substantive code with an incomplete key mechanism or data flow & The real path was run and left credible partial evidence, but the key path is unfinished & Real and admissible but mixed or weak support & Substantive work lacking a key protocol element, comparator, or piece of evidence\\\midrule
3 & Core implementation correct, with one limited gap in interface, tests, or integration & Run completed, with one limited protocol or evidence gap & Credible result that mostly meets the criterion, with the unmet parts stated & Complete, adjudicable execution whose result misses the criterion\\\midrule
4 & Complete and correct implementation, with enough code or tests to support the judgment & Qualified run with correct protocol and controls, and with its commands and logs & Admissible result that meets the criterion with correctly bounded scope & Qualified execution whose admissible evidence meets the criterion\\
\bottomrule
\end{tabularx}
\end{table}

\paragraph{Claim verdicts.}
The judge maps the evaluated system's own records onto the ten required claims, since these records may use different claim names or combine several required claims in one experiment. A claim is \emph{met} only when admissible evidence satisfies its frozen criterion. It is \emph{unmet} when the criterion fails, when a required element such as the comparator is missing, or when no corresponding evidence exists. A run cannot leave a claim unresolved. Only an evaluator failure yields an error, and it blocks the execution score instead of counting against the system.

\paragraph{Integrity gates.}
Table~\ref{tab:gates} lists the eight gates. A gate is \emph{triggered} only by independently verifiable evidence, and it then sets the rubric items that depend on the contaminated evidence to zero. The first four gates in Table~\ref{tab:gates} are the most severe. If one of them contaminates evidence whose extent cannot be bounded, the whole workflow is invalidated and scores zero. A suspicion without such evidence is recorded as a review flag, and a finding whose evidence is incomplete is recorded as indeterminate. Neither changes a score. Across the stage evaluations of all scored workflows (Appendix~\ref{app:coverage}), one gate was triggered. In the end-to-end run of Codex with GPT-5.5, the paper's only figure was plotted from a different result file than its caption and source state. It therefore contradicts the table on the same page, and the two paper-writing items on result lineage and render consistency were set to zero. All other findings were review flags or indeterminate.

{\footnotesize
\setlength{\LTleft}{0pt}\setlength{\LTright}{0pt}
\setlength{\LTpre}{20pt}\setlength{\LTpost}{0pt}
\begin{longtable}{@{}l>{\raggedright\arraybackslash}p{.72\linewidth}@{}}
\caption{Integrity gates and their triggers. Each gate zeros only the rubric items that depend on the affected evidence, unless the contamination cannot be bounded.}\label{tab:gates}\\
\toprule Gate & Trigger\\\midrule
\endfirsthead
\toprule Gate & Trigger\\\midrule
\endhead
Forbidden source & Use of the anchor paper with its supplementary material, code, or checkpoints, or use of sources released after the cutoff\\\midrule
Reference leak & Reference outputs or rubric contents enter the workspace or the submission, or so does the mapping from reference entries to paper locations\\\midrule
Test protocol & Unlogged test access, behavior labels in a calibration or input described as unlabeled, or optimized test results described as untouched\\\midrule
Fabricated execution & An unexecuted experiment or a paper-reported number presented as a local result\\\midrule
Session mismatch & Subject, session, or sample mapping inconsistent with the frozen split manifest\\\midrule
Unsupported claim & A claim that exceeds its evidence or depends on a failed, missing, or invalid experiment\\\midrule
Result inconsistency & One result reported with unexplained conflicts across result files, figures, and text\\\midrule
Paper claim expansion & The paper broadens the frozen claims or contradicts the analyzed outcomes and limitations\\
\bottomrule
\end{longtable}
}

\FloatBarrier
\section{Evaluated runs, system configuration, and resource accounting}\label{app:coverage}
Table~\ref{tab:runs} lists every scored workflow with its overview score and resource use. The runs of GPT-5.6 Sol and GPT-6 Astra were executed in a later batch. The skill-augmented harnesses ran on a second host and the research-orchestration systems on a third, under the same task, budget, and scoring.

\paragraph{System configuration.}
Table~\ref{tab:config} lists the version, model, and settings of every configuration. Each value is read from the run's capability and invocation records. The research-orchestration systems record no capability receipt, so their values come from their build configuration and retained session records. The stagewise runs use the same configuration as the corresponding end-to-end runs.

\begin{table}[ht]
\caption{Scored workflows, overview scores, and resource use. Wall-clock time covers the whole workflow. For stagewise runs it sums the seven stage runs and excludes queueing. Token counts include cached input. The count of AutoResearchClaw is a lower bound because its backend records do not cover the whole run, and n/a marks the stagewise run of Kimi Code, whose receipts record no token use.}\label{tab:runs}
\centering\footnotesize\input{tables/tabA_runs_resources}
\end{table}

\begin{table}[ht]
\caption{Configuration of the evaluated systems. All runs request high reasoning effort and allow no fallback model. Context is the model context window. GPT-5.6 Sol and GPT-6 Astra set it explicitly to 872K tokens. The GPT-5.5 runs set no window, so Codex uses its default of 272K tokens for this model and makes 95\% (258.4K) of it usable, as the retained sessions of ARIS, Orchestra, and DeepScientist record. Claude Code also calls \texttt{claude-haiku-4-5} as an auxiliary model.}\label{tab:config}
\centering\footnotesize
\begin{tabularx}{\linewidth}{@{}>{\raggedright\arraybackslash}X>{\raggedright\arraybackslash}p{.3\linewidth}lcr@{}}
\toprule System & Harness or system version & Model & Effort & Context\\\midrule
Codex / GPT-5.5 & Codex CLI 0.153.4 & \texttt{gpt-5.5} & high & 272K\\\midrule
Codex / GPT-5.6 Sol & Codex CLI 0.153.4 & \texttt{gpt-5.6-sol} & high & 872K\\\midrule
Codex / GPT-6 Astra & Codex CLI 0.153.4 & \texttt{gpt-6-astra} & high & 872K\\\midrule
Claude Code / Opus 4.8 & Claude Code 2.1.269 & \texttt{claude-opus-4-8} & high & 1M\\\midrule
Kimi Code / K3 & Kimi Code 0.39.1 & \texttt{k3} & high & 1M\\\midrule
ARIS & Codex CLI 0.153.4 with ARIS skills (commit \texttt{94d8093}) & \texttt{gpt-5.5} & high & 272K\\\midrule
Orchestra & Codex CLI 0.153.4 with Orchestra skills (commit \texttt{773a529}) & \texttt{gpt-5.5} & high & 272K\\\midrule
AutoResearchClaw & AutoResearchClaw 0.5.0 (commit \texttt{be4ba47}) & \texttt{gpt-5.5} & high & 272K\\\midrule
DeepScientist & DeepScientist 1.6.0 (commit \texttt{b366244}) on Codex CLI 0.153.4 & \texttt{gpt-5.5} & high & 272K\\
\bottomrule
\end{tabularx}
\end{table}

\FloatBarrier
\begingroup
\setlength{\textfloatsep}{8pt}
\setlength{\floatsep}{8pt}
\section{Stagewise scores and per-claim verdicts}\label{app:claims}
Table~\ref{tab:stagewise-scores} lists the stage and dimension scores behind Figure~\ref{fig:stages}. Table~\ref{tab:claim-matrix} gives the judge's ratings for every required claim in every run and underlies the claim counts in Sections~\ref{sec:res-e2e} and~\ref{sec:res-exec}. Of the 90 end-to-end claims, 14 are met, and E5 for GPT-6 Astra is the only empirical claim met in an end-to-end run. Both tables show the first run under the original judge, as submitted. Appendix~\ref{app:repeat} gives both runs under both judges.

\begin{table}[ht]
\caption{Stage scores, execution-dimension scores, and overview scores of the three general-purpose harnesses in end-to-end and stagewise evaluation. Mod., Code, Qual., and Comp. are module development, code execution, result quality, and claim completion.}\label{tab:stagewise-scores}
\centering\footnotesize\input{tables/tabA_stagewise_scores}
\end{table}

\begin{table}[ht]
\caption{Per-claim ratings as claim completion/result quality on the 0--4 scale. Shaded bold cells mark claims whose success criterion is met. Met is the number of claims met. Rep. is the number of claims the evaluated system reported as attained by the end of execution. Att. is the number of attempts on required claims. Reported claims and attempts are not tracked for the stagewise execution runs.}\label{tab:claim-matrix}
\centering\footnotesize\input{tables/tabA_claim_matrix}
\end{table}

\paragraph{Example of reported success without a met criterion.}
GPT-5.6 Sol reported eight attained claims, but only MC1--MC3 meet their criteria. The judge's rationales for the unmet claims show where its evidence departs from the criteria. For E1, the system compared its decoder with baselines of its own choosing instead of the frozen Wiener filter, RNN, and calibration-aware peers, and its decoder still trailed one of its own baselines on H1. For E2, one-shot calibration on M1 came within 0.05 of full calibration. One-shot $R^2$ on M2 and H1 was negative and below a trivial predictor, however, so the all-subject requirement fails. For E4, the system tested a single 70\% retention level, so the required 20\% condition does not exist. For E5, it timed only H1-shaped inputs and reported absolute latency instead of the required ratio. The system's records organize its experiments under its own claim names and scopes, so a reported success can refer to a narrower experiment than the required claim. These are the gaps that separate the reported and confirmed tracks in Figures~\ref{fig:iteration}(a) and~\ref{fig:iteration}(b).

\FloatBarrier
\endgroup

\section{Repeated runs and a second judge}\label{app:repeat}
\paragraph{Runs and judges.} Every configuration has two research runs, each scored by \texttt{claude-opus-5}, the original judge, and by \texttt{gpt-6-astra} (reasoning effort \texttt{xhigh}), a non-Anthropic model, with the same rubrics, evidence, and frozen criteria. Table~\ref{tab:landscape} and Figure~\ref{fig:stages} report the mean of the four evaluations of each configuration, and the tables below report each one.

\paragraph{What holds.} Table~\ref{tab:rebuttal-summary} summarizes the nine end-to-end systems under each evaluation. Design and planning exceeds execution on average in every evaluation, and for every system on the mean of four. With nine systems as units, an exact two-sided Wilcoxon signed-rank test gives $p=0.004$. Stagewise runs score higher than end-to-end runs for all three harnesses in every evaluation (Table~\ref{tab:rebuttal-stagewise}). The two judges rank the systems similarly (Spearman 0.98 between their run means).

\begin{table}[ht]
\caption{Nine end-to-end systems under each evaluation. Research-part and overview scores are means over the systems. Design $-$ exec.\ is the mean difference between design and planning and experimental execution, and Design $>$ exec.\ counts the systems where it is positive. Opus 5 and Astra are \texttt{claude-opus-5} and \texttt{gpt-6-astra}. r1* is the submitted evaluation.}\label{tab:rebuttal-summary}
\centering\footnotesize\input{tables/tabA_rebuttal_summary}
\end{table}

\FloatBarrier
{\footnotesize
\setlength{\LTleft}{0pt}\setlength{\LTright}{0pt}
\setlength{\LTpre}{20pt}\setlength{\LTpost}{0pt}
\input{tables/tabA_rebuttal_stagewise}
}

\FloatBarrier
{\footnotesize
\setlength{\LTleft}{0pt}\setlength{\LTright}{0pt}
\setlength{\LTpre}{20pt}\setlength{\LTpost}{0pt}
\input{tables/tabA_rebuttal_full}
}

\FloatBarrier
\section{A second anchor paper}\label{app:fits}
\paragraph{Task.} We built a second task from FITS, \emph{Modeling Time Series with 10k Parameters} (ICLR 2024), a compact frequency-domain model for long-horizon time-series forecasting and reconstruction-based anomaly detection. The construction, stage protocol, rubric structure, and judge contract are those of Section~\ref{sec:framework}. The task has eleven required claims and 93 rubric items. Eight end-to-end systems and three stagewise harnesses each ran it twice, with the GPT-5.5 conditions of the first task replaced by GPT-5.6 Sol.

\paragraph{Scores.} Table~\ref{tab:fits-r1} gives the first run of all eleven workflows under \texttt{claude-opus-5}. Stagewise runs again score above end-to-end runs for all three harnesses run in both modes, by 9.0, 33.5, and 33.4 overview points. Experimental execution again scores below design and planning on average over the eight end-to-end systems (41.8 against 66.9). End-to-end attainment of the required claims also remains low, at 25 of the 88 required claims.

\begin{table}[!ht]
\caption{First run of the eleven FITS workflows under \texttt{claude-opus-5}. Columns as in Table~\ref{tab:landscape}. Claims met is out of the eleven required claims. Single run, single judge.}\label{tab:fits-r1}
\centering\scriptsize\input{tables/tabA_t1_fits_r1}
\end{table}

%% file: tables/tabA_rubric_structure.tex
{\setlength{\tabcolsep}{4pt}%
\begin{longtable}{@{}l>{\raggedright\arraybackslash}p{.42\linewidth}rrr@{}}
\caption{Rubric structure of the task, with its dimensions, item counts, and weights. The literature-review dimensions each contain items for the sources found, the comparison with prior work, and the resulting gap. The experimental protocol of experiment planning covers the data, metrics, baselines, and training. The two paper-writing dimensions are scored in separate judge calls.}\label{tab:rubric-structure}\\
\toprule
Stage & Dimension & Items & \shortstack{Weight in\\stage} & \shortstack{Stage weight\\in part}\\
\midrule
\endfirsthead
\toprule
Stage & Dimension & Items & \shortstack{Weight in\\stage} & \shortstack{Stage weight\\in part}\\
\midrule
\endhead
\multicolumn{5}{@{}l}{\textbf{Design and planning} (30\% of the overview score)}\\
S1 Literature review & Intracortical motor decoding & 3 & 25\% & 20\%\\
\cmidrule(l){2-4}
 & Cross-session neural alignment & 3 & 25\% & \\
\cmidrule(l){2-4}
 & Neuroscience foundation models & 3 & 25\% & \\
\cmidrule(l){2-4}
 & Set-structured (permutation-invariant) models & 3 & 25\% & \\
\midrule
S2 Research claims & Method claims MC1--MC4 & 4 & 40\% & 20\%\\
\cmidrule(l){2-4}
 & Empirical claims E1--E6 & 6 & 60\% & \\
\midrule
S3 Method design & Method specification (eight components) & 8 & 100\% & 30\%\\
\midrule
S4 Experiment planning & Experimental protocol & 4 & 33\% & 30\%\\
\cmidrule(l){2-4}
 & Core experiments (eight) & 8 & 67\% & \\
\midrule
\multicolumn{5}{@{}l}{\textbf{Experimental execution} (50\% of the overview score)}\\
S5 Experimental execution & Module development & 7 & 25\% & 100\%\\
\cmidrule(l){2-4}
 & Code execution & 10 & 25\% & \\
\cmidrule(l){2-4}
 & Result quality (one item per claim) & 10 & 25\% & \\
\cmidrule(l){2-4}
 & Claim completion (one item per claim) & 10 & 25\% & \\
\midrule
\multicolumn{5}{@{}l}{\textbf{Analysis and writing} (20\% of the overview score)}\\*
S6 Results analysis & Per-claim outcome and analysis & 10 & 100\% & 40\%\\
\midrule
S7 Paper writing & Scientific writing and traceability & 5 & 50\% & 60\%\\
\cmidrule(l){2-4}
 & Conference review of the rendered paper & 5 & 50\% & \\
\midrule
\multicolumn{2}{@{}l}{Total} & 99 & & \\
\bottomrule
\end{longtable}}

%% file: tables/tabA_runs_resources.tex
{\setlength{\tabcolsep}{4pt}%
\begin{tabularx}{\linewidth}{@{}>{\raggedright\arraybackslash}Xlrrr@{}}
\toprule
System & Mode & Overview & Wall-clock (h) & Tokens (M)\\
\midrule
Codex / GPT-5.5 & End to end & 45.1 & 0.30 & 4.7\\
\midrule
Codex / GPT-5.6 Sol & End to end & 59.5 & 1.72 & 57.2\\
\midrule
Codex / GPT-6 Astra & End to end & 64.9 & 1.10 & 12.8\\
\midrule
Claude Code / Opus 4.8 & End to end & 61.3 & 5.73 & 126.5\\
\midrule
Kimi Code / K3 & End to end & 64.2 & 20.56 & 92.0\\
\midrule
ARIS & End to end & 44.5 & 0.74 & 15.1\\
\midrule
Orchestra & End to end & 45.2 & 0.90 & 19.2\\
\midrule
AutoResearchClaw & End to end & 18.6 & 0.71 & 0.8\\
\midrule
DeepScientist & End to end & 39.9 & 7.00 & 205.7\\
\midrule
Codex / GPT-5.5 & Stagewise & 78.5 & 1.21 & 15.0\\
\midrule
Claude Code / Opus 4.8 & Stagewise & 81.0 & 3.58 & 69.3\\
\midrule
Kimi Code / K3 & Stagewise & 74.8 & 40.56 & n/a\\
\bottomrule
\end{tabularx}}

%% file: tables/tabA_stagewise_scores.tex
{\setlength{\tabcolsep}{1.5pt}%
\begin{tabularx}{\linewidth}{@{}>{\raggedright\arraybackslash}Xl*{12}{c}@{}}
\toprule
 & & \multicolumn{7}{c}{Stage} & \multicolumn{4}{c}{S5 dimension} & \\
\cmidrule(lr){3-9}\cmidrule(lr){10-13}
System & Mode & S1 & S2 & S3 & S4 & S5 & S6 & S7 & Mod. & Code & Qual. & Comp. & Overview\\
\midrule
Codex / GPT-5.5 & End to end & 68.8 & 47.5 & 56.2 & 41.7 & 44.0 & 42.5 & 32.5 & 53.6 & 50.0 & 30.0 & 42.5 & 45.1\\
\cmidrule(l){2-14}
 & Stagewise & 83.3 & 67.5 & 90.6 & 83.3 & 77.6 & 90.0 & 65.0 & 67.9 & 65.0 & 85.0 & 92.5 & 78.5\\
\midrule
Claude Code / Opus 4.8 & End to end & 81.2 & 65.0 & 81.2 & 66.7 & 57.2 & 50.0 & 55.0 & 71.4 & 57.5 & 42.5 & 57.5 & 61.3\\
\cmidrule(l){2-14}
 & Stagewise & 89.6 & 72.5 & 90.6 & 70.8 & 86.1 & 77.5 & 62.5 & 89.3 & 67.5 & 92.5 & 95.0 & 81.0\\
\midrule
Kimi Code / K3 & End to end & 72.9 & 65.0 & 78.1 & 68.8 & 61.6 & 62.5 & 57.5 & 71.4 & 62.5 & 47.5 & 65.0 & 64.2\\
\cmidrule(l){2-14}
 & Stagewise & 81.2 & 77.5 & 81.2 & 72.9 & 74.9 & 80.0 & 62.5 & 82.1 & 65.0 & 75.0 & 77.5 & 74.8\\
\bottomrule
\end{tabularx}}

%% file: tables/tabA_claim_matrix.tex
{\setlength{\tabcolsep}{2.4pt}\renewcommand{\arraystretch}{1.05}%
\begin{tabularx}{\linewidth}{@{}>{\raggedright\arraybackslash}X*{10}{c}ccc@{}}
\toprule
System & MC1 & MC2 & MC3 & MC4 & E1 & E2 & E3 & E4 & E5 & E6 & Met & Rep. & Att.\\
\midrule
\multicolumn{14}{@{}l}{\textit{General-purpose agent harness, end to end}}\\
\hspace{0.5em}Codex / GPT-5.5 & \cellcolor{black!12}\textbf{4/4} & 2/3 & 2/2 & 0/0 & 2/0 & 2/1 & 0/0 & 1/0 & 2/2 & 2/0 & 1 & 0 & 9\\
\midrule
\hspace{0.5em}Codex / GPT-5.6 Sol & \cellcolor{black!12}\textbf{4/4} & \cellcolor{black!12}\textbf{4/4} & \cellcolor{black!12}\textbf{4/4} & 1/0 & 2/0 & 3/1 & 0/0 & 2/0 & 2/0 & 2/0 & 3 & 8 & 24\\
\midrule
\hspace{0.5em}Codex / GPT-6 Astra & \cellcolor{black!12}\textbf{4/4} & \cellcolor{black!12}\textbf{4/4} & \cellcolor{black!12}\textbf{4/4} & 0/0 & 3/1 & 3/2 & 0/0 & 2/0 & \cellcolor{black!12}\textbf{4/4} & 2/0 & 4 & 8 & 10\\
\midrule
\hspace{0.5em}Claude Code / Opus 4.8 & \cellcolor{black!12}\textbf{4/4} & 2/3 & 2/2 & 2/2 & 3/1 & 3/1 & 0/0 & 2/0 & 2/2 & 3/2 & 1 & 3 & 9\\
\midrule
\hspace{0.5em}Kimi Code / K3 & \cellcolor{black!12}\textbf{4/4} & \cellcolor{black!12}\textbf{4/4} & \cellcolor{black!12}\textbf{4/4} & 2/2 & 2/0 & 3/2 & 0/0 & 2/0 & 2/2 & 3/1 & 3 & 2 & 18\\
\midrule
\multicolumn{14}{@{}l}{\textit{Skill-augmented harness (Codex / GPT-5.5), end to end}}\\
\hspace{0.5em}ARIS & 2/2 & \cellcolor{black!12}\textbf{4/4} & 1/0 & 0/0 & 2/0 & 2/2 & 0/0 & 1/0 & 1/0 & 0/0 & 1 & 8 & 28\\
\midrule
\hspace{0.5em}Orchestra & 2/1 & \cellcolor{black!12}\textbf{4/4} & 2/2 & 0/0 & 2/0 & 2/1 & 0/0 & 1/0 & 1/0 & 1/0 & 1 & 8 & 10\\
\midrule
\multicolumn{14}{@{}l}{\textit{Research-orchestration system (GPT-5.5), end to end}}\\
\hspace{0.5em}AutoResearchClaw & 0/0 & 0/0 & 0/0 & 0/0 & 0/0 & 0/0 & 0/0 & 0/0 & 0/0 & 0/0 & 0 & 0 & 0\\
\midrule
\hspace{0.5em}DeepScientist & 1/0 & 0/0 & 0/0 & 0/0 & 2/0 & 0/0 & 0/0 & 0/0 & 0/0 & 0/0 & 0 & 0 & 34\\
\midrule
\multicolumn{14}{@{}l}{\textit{General-purpose agent harness, stagewise}}\\
\midrule
\hspace{0.5em}Codex / GPT-5.5 & \cellcolor{black!12}\textbf{4/4} & \cellcolor{black!12}\textbf{4/4} & \cellcolor{black!12}\textbf{4/4} & \cellcolor{black!12}\textbf{4/4} & \cellcolor{black!12}\textbf{4/4} & 3/2 & 3/2 & \cellcolor{black!12}\textbf{4/4} & \cellcolor{black!12}\textbf{4/4} & 3/2 & 7 & -- & --\\
\midrule
\hspace{0.5em}Claude Code / Opus 4.8 & \cellcolor{black!12}\textbf{4/4} & \cellcolor{black!12}\textbf{4/4} & \cellcolor{black!12}\textbf{4/4} & \cellcolor{black!12}\textbf{4/4} & \cellcolor{black!12}\textbf{4/4} & \cellcolor{black!12}\textbf{4/4} & \cellcolor{black!12}\textbf{4/4} & 3/3 & \cellcolor{black!12}\textbf{4/4} & 3/2 & 8 & -- & --\\
\midrule
\hspace{0.5em}Kimi Code / K3 & \cellcolor{black!12}\textbf{4/4} & \cellcolor{black!12}\textbf{4/4} & \cellcolor{black!12}\textbf{4/4} & \cellcolor{black!12}\textbf{4/4} & \cellcolor{black!12}\textbf{4/4} & \cellcolor{black!12}\textbf{4/4} & \cellcolor{black!12}\textbf{4/4} & 3/2 & 0/0 & 0/0 & 7 & -- & --\\
\bottomrule
\end{tabularx}}

%% file: tables/tabA_rebuttal_summary.tex
{\setlength{\tabcolsep}{4pt}%
\begin{tabular}{@{}lccccccc@{}}
\toprule
Judge, run & \shortstack{Design\\\& plan.} & \shortstack{Exec-\\ution} & \shortstack{Anal.\\\& writ.} & Overview & \shortstack{Design\\$-$ exec.} & \shortstack{Design\\$>$ exec.} & \shortstack{Claims\\met} \\
\midrule
Opus 5, r1* & 61.6 & 43.0 & 46.5 & 49.3 & 18.7 & 9/9 & 14/90 \\
Opus 5, r2 & 65.6 & 51.0 & 54.1 & 56.0 & 14.5 & 8/9 & 16/90 \\
Astra, r1 & 58.1 & 38.4 & 41.6 & 44.9 & 19.7 & 8/9 & 5/90 \\
Astra, r2 & 58.2 & 36.3 & 40.8 & 43.8 & 21.9 & 9/9 & 9/90 \\
\midrule
\textbf{Mean of the four} & 60.9 & 42.2 & 45.7 & 48.5 & 18.7 & 9/9 & -- \\
\bottomrule
\end{tabular}}

%% file: tables/tabA_rebuttal_stagewise.tex
{\setlength{\tabcolsep}{3pt}%
\begin{longtable}{@{}ll*{7}{c}cc@{}}
\caption{Stage and overview scores of the three general-purpose harnesses in stagewise and end-to-end runs, for each run and judge. Met is the number of required claims met (of 10). Judges and r1* as in Table~\ref{tab:rebuttal-summary}. Lit., Meth., Exec., Anal., and Writ.\ are literature review, method design, experimental execution, results analysis, and paper writing.}\label{tab:rebuttal-stagewise}\\
\toprule
Mode & Judge, run & Lit. & Claims & Meth. & Plan & Exec. & Anal. & Writ. & Overview & Met \\
\midrule
\endfirsthead
\toprule
Mode & Judge, run & Lit. & Claims & Meth. & Plan & Exec. & Anal. & Writ. & Overview & Met \\
\midrule
\endhead
\bottomrule
\endfoot
\multicolumn{11}{@{}l}{\textbf{Codex / GPT-5.5}}\\*
\hspace{0.7em}Stagewise & Opus 5, r1* & 83.3 & 67.5 & 90.6 & 83.3 & 77.6 & 90.0 & 65.0 & 78.5 & 7 \\*
 & Opus 5, r2 & 79.2 & 67.5 & 93.8 & 83.3 & 48.1 & 87.5 & 57.5 & 62.7 & 3 \\*
 & Astra, r1 & 83.3 & 50.0 & 90.6 & 62.5 & 61.6 & 70.0 & 52.5 & 64.5 & 2 \\*
 & Astra, r2 & 79.2 & 52.5 & 87.5 & 62.5 & 52.9 & 72.5 & 47.5 & 59.3 & 4 \\*[4pt]
\hspace{0.7em}End-to-end & Opus 5, r1* & 68.8 & 47.5 & 56.2 & 41.7 & 44.0 & 42.5 & 32.5 & 45.1 & 1 \\*
 & Opus 5, r2 & 62.5 & 50.0 & 59.4 & 56.2 & 46.5 & 42.5 & 42.5 & 48.9 & 3 \\*
 & Astra, r1 & 75.0 & 52.5 & 59.4 & 45.8 & 36.9 & 45.0 & 32.5 & 43.1 & 0 \\*
 & Astra, r2 & 66.7 & 45.0 & 68.8 & 43.8 & 34.6 & 45.0 & 42.5 & 42.8 & 0 \\
\midrule
\multicolumn{11}{@{}l}{\textbf{Claude Code / Opus 4.8}}\\*
\hspace{0.7em}Stagewise & Opus 5, r1* & 89.6 & 72.5 & 90.6 & 70.8 & 86.1 & 77.5 & 62.5 & 81.0 & 8 \\*
 & Opus 5, r2 & 83.3 & 62.5 & 90.6 & 68.8 & 84.6 & 87.5 & 62.5 & 79.9 & 8 \\*
 & Astra, r1 & 70.8 & 50.0 & 84.4 & 58.3 & 80.1 & 7.5 & 25.0 & 63.7 & 6 \\*
 & Astra, r2 & 75.0 & 47.5 & 81.2 & 56.2 & 84.8 & 67.5 & 27.5 & 70.8 & 7 \\*[4pt]
\hspace{0.7em}End-to-end & Opus 5, r1* & 81.2 & 65.0 & 81.2 & 66.7 & 57.2 & 50.0 & 55.0 & 61.3 & 1 \\*
 & Opus 5, r2 & 87.5 & 65.0 & 96.9 & 79.2 & 75.0 & 42.5 & 75.0 & 74.9 & 3 \\*
 & Astra, r1 & 66.7 & 42.5 & 78.1 & 58.3 & 47.0 & 55.0 & 35.0 & 50.9 & 2 \\*
 & Astra, r2 & 70.8 & 37.5 & 71.9 & 45.8 & 30.9 & 42.5 & 47.5 & 41.6 & 0 \\
\midrule
\multicolumn{11}{@{}l}{\textbf{Kimi Code / K3}}\\*
\hspace{0.7em}Stagewise & Opus 5, r1* & 81.2 & 77.5 & 81.2 & 72.9 & 74.9 & 80.0 & 62.5 & 74.8 & 7 \\*
 & Opus 5, r2 & 85.4 & 82.5 & 90.6 & 72.9 & 85.2 & 87.5 & 55.0 & 81.0 & 7 \\*
 & Astra, r1 & 54.2 & 60.0 & 78.1 & 50.0 & 68.0 & 70.0 & 37.5 & 62.5 & 4 \\*
 & Astra, r2 & 64.6 & 60.0 & 93.8 & 56.2 & 79.0 & 45.0 & 35.0 & 68.3 & 7 \\*[4pt]
\hspace{0.7em}End-to-end & Opus 5, r1* & 72.9 & 65.0 & 78.1 & 68.8 & 61.6 & 62.5 & 57.5 & 64.2 & 3 \\*
 & Opus 5, r2 & 91.7 & 97.5 & 93.8 & 43.8 & 65.0 & 87.5 & 82.5 & 73.1 & 4 \\*
 & Astra, r1 & 62.5 & 55.0 & 43.8 & 18.8 & 51.3 & 50.0 & 37.5 & 46.8 & 3 \\*
 & Astra, r2 & 72.9 & 60.0 & 71.9 & 52.1 & 53.6 & 60.0 & 25.0 & 53.7 & 3 \\
\end{longtable}}

%% file: tables/tabA_rebuttal_full.tex
{\setlength{\tabcolsep}{3pt}%
\begin{longtable}{@{}l*{11}{c}c@{}}
\caption{End-to-end scores of every run under each judge. Columns as in Table~\ref{tab:landscape}. Met is the number of required claims met (of 10). Opus 5 and Astra are \texttt{claude-opus-5} and \texttt{gpt-6-astra}. r1* is the submitted evaluation.}\label{tab:rebuttal-full}\\
\toprule
Judge, run & Lit. & Claims & Meth. & Plan & Module & Code & Result & Claim & Anal. & Writ. & Overview & Met \\
\midrule
\endfirsthead
\toprule
Judge, run & Lit. & Claims & Meth. & Plan & Module & Code & Result & Claim & Anal. & Writ. & Overview & Met \\
\midrule
\endhead
\bottomrule
\endfoot
\multicolumn{13}{@{}l}{\textbf{Codex / GPT-5.5}}\\*
\hspace{0.7em}Opus 5, r1* & 68.8 & 47.5 & 56.2 & 41.7 & 53.6 & 50.0 & 30.0 & 42.5 & 42.5 & 32.5 & 45.1 & 1 \\*
\hspace{0.7em}Opus 5, r2 & 62.5 & 50.0 & 59.4 & 56.2 & 53.6 & 47.5 & 35.0 & 50.0 & 42.5 & 42.5 & 48.9 & 3 \\*
\hspace{0.7em}Astra, r1 & 75.0 & 52.5 & 59.4 & 45.8 & 50.0 & 42.5 & 15.0 & 40.0 & 45.0 & 32.5 & 43.1 & 0 \\*
\hspace{0.7em}Astra, r2 & 66.7 & 45.0 & 68.8 & 43.8 & 53.6 & 27.5 & 22.5 & 35.0 & 45.0 & 42.5 & 42.8 & 0 \\
\midrule
\multicolumn{13}{@{}l}{\textbf{Codex / GPT-5.6 Sol}}\\*
\hspace{0.7em}Opus 5, r1* & 83.3 & 65.0 & 78.1 & 66.7 & 60.7 & 52.5 & 32.5 & 60.0 & 55.0 & 62.5 & 59.5 & 3 \\*
\hspace{0.7em}Opus 5, r2 & 83.3 & 87.5 & 96.9 & 91.7 & 64.3 & 42.5 & 100.0 & 100.0 & 85.0 & 80.0 & 82.0 & 3 \\*
\hspace{0.7em}Astra, r1 & 87.5 & 52.5 & 81.2 & 56.2 & 67.9 & 52.5 & 22.5 & 45.0 & 60.0 & 60.0 & 56.3 & 0 \\*
\hspace{0.7em}Astra, r2 & 87.5 & 52.5 & 78.1 & 54.2 & 67.9 & 45.0 & 30.0 & 55.0 & 50.0 & 57.5 & 55.9 & 3 \\
\midrule
\multicolumn{13}{@{}l}{\textbf{Codex / GPT-6 Astra}}\\*
\hspace{0.7em}Opus 5, r1* & 72.9 & 65.0 & 84.4 & 70.8 & 64.3 & 57.5 & 47.5 & 65.0 & 62.5 & 70.0 & 64.9 & 4 \\*
\hspace{0.7em}Opus 5, r2 & 70.8 & 77.5 & 93.8 & 93.8 & 96.4 & 100.0 & 100.0 & 100.0 & 92.5 & 80.0 & 92.3 & 3 \\*
\hspace{0.7em}Astra, r1 & 93.8 & 65.0 & 87.5 & 62.5 & 67.9 & 60.0 & 27.5 & 47.5 & 62.5 & 67.5 & 61.5 & 0 \\*
\hspace{0.7em}Astra, r2 & 85.4 & 57.5 & 87.5 & 64.6 & 71.4 & 57.5 & 30.0 & 57.5 & 65.0 & 75.0 & 63.5 & 3 \\
\midrule
\multicolumn{13}{@{}l}{\textbf{Claude Code / Opus 4.8}}\\*
\hspace{0.7em}Opus 5, r1* & 81.2 & 65.0 & 81.2 & 66.7 & 71.4 & 57.5 & 42.5 & 57.5 & 50.0 & 55.0 & 61.3 & 1 \\*
\hspace{0.7em}Opus 5, r2 & 87.5 & 65.0 & 96.9 & 79.2 & 75.0 & 45.0 & 85.0 & 95.0 & 42.5 & 75.0 & 74.9 & 3 \\*
\hspace{0.7em}Astra, r1 & 66.7 & 42.5 & 78.1 & 58.3 & 67.9 & 40.0 & 25.0 & 55.0 & 55.0 & 35.0 & 50.9 & 2 \\*
\hspace{0.7em}Astra, r2 & 70.8 & 37.5 & 71.9 & 45.8 & 53.6 & 12.5 & 20.0 & 37.5 & 42.5 & 47.5 & 41.6 & 0 \\
\midrule
\multicolumn{13}{@{}l}{\textbf{Kimi Code / K3}}\\*
\hspace{0.7em}Opus 5, r1* & 72.9 & 65.0 & 78.1 & 68.8 & 71.4 & 62.5 & 47.5 & 65.0 & 62.5 & 57.5 & 64.2 & 3 \\*
\hspace{0.7em}Opus 5, r2 & 91.7 & 97.5 & 93.8 & 43.8 & 75.0 & 67.5 & 57.5 & 60.0 & 87.5 & 82.5 & 73.1 & 4 \\*
\hspace{0.7em}Astra, r1 & 62.5 & 55.0 & 43.8 & 18.8 & 67.9 & 47.5 & 30.0 & 60.0 & 50.0 & 37.5 & 46.8 & 3 \\*
\hspace{0.7em}Astra, r2 & 72.9 & 60.0 & 71.9 & 52.1 & 64.3 & 52.5 & 35.0 & 62.5 & 60.0 & 25.0 & 53.7 & 3 \\
\midrule
\multicolumn{13}{@{}l}{\textbf{ARIS}}\\*
\hspace{0.7em}Opus 5, r1* & 66.7 & 52.5 & 56.2 & 41.7 & 53.6 & 45.0 & 20.0 & 32.5 & 42.5 & 52.5 & 44.5 & 1 \\*
\hspace{0.7em}Opus 5, r2 & 75.0 & 45.0 & 56.2 & 41.7 & 42.9 & 40.0 & 25.0 & 40.0 & 45.0 & 57.5 & 45.0 & 0 \\*
\hspace{0.7em}Astra, r1 & 72.9 & 45.0 & 62.5 & 41.7 & 50.0 & 42.5 & 17.5 & 40.0 & 47.5 & 35.0 & 43.2 & 0 \\*
\hspace{0.7em}Astra, r2 & 79.2 & 37.5 & 59.4 & 45.8 & 42.9 & 37.5 & 15.0 & 37.5 & 32.5 & 30.0 & 39.3 & 0 \\
\midrule
\multicolumn{13}{@{}l}{\textbf{Orchestra}}\\*
\hspace{0.7em}Opus 5, r1* & 66.7 & 52.5 & 53.1 & 45.8 & 50.0 & 50.0 & 20.0 & 37.5 & 47.5 & 47.5 & 45.2 & 1 \\*
\hspace{0.7em}Opus 5, r2 & 58.3 & 50.0 & 40.6 & 47.9 & 21.4 & 17.5 & 10.0 & 25.0 & 30.0 & 35.0 & 30.3 & 0 \\*
\hspace{0.7em}Astra, r1 & 77.1 & 45.0 & 59.4 & 50.0 & 50.0 & 47.5 & 17.5 & 42.5 & 45.0 & 42.5 & 45.6 & 0 \\*
\hspace{0.7em}Astra, r2 & 64.6 & 45.0 & 43.8 & 45.8 & 39.3 & 22.5 & 5.0 & 37.5 & 42.5 & 37.5 & 35.6 & 0 \\
\midrule
\multicolumn{13}{@{}l}{\textbf{AutoResearchClaw}}\\*
\hspace{0.7em}Opus 5, r1* & 22.9 & 40.0 & 62.5 & 18.8 & 42.9 & 17.5 & 0.0 & 0.0 & 0.0 & 0.0 & 18.6 & 0 \\*
\hspace{0.7em}Opus 5, r2 & 47.9 & 0.0 & 59.4 & 27.1 & 50.0 & 12.5 & 0.0 & 12.5 & 0.0 & 0.0 & 20.0 & 0 \\*
\hspace{0.7em}Astra, r1 & 27.1 & 40.0 & 68.8 & 27.1 & 46.4 & 22.5 & 0.0 & 0.0 & 0.0 & 0.0 & 21.3 & 0 \\*
\hspace{0.7em}Astra, r2 & 39.6 & 35.0 & 71.9 & 37.5 & 57.1 & 20.0 & 0.0 & 12.5 & 0.0 & 0.0 & 25.5 & 0 \\
\midrule
\multicolumn{13}{@{}l}{\textbf{DeepScientist}}\\*
\hspace{0.7em}Opus 5, r1* & 70.8 & 57.5 & 75.0 & 58.3 & 53.6 & 25.0 & 0.0 & 7.5 & 42.5 & 50.0 & 39.9 & 0 \\*
\hspace{0.7em}Opus 5, r2 & 64.6 & 37.5 & 68.8 & 52.1 & 46.4 & 32.5 & 0.0 & 12.5 & 37.5 & 50.0 & 37.4 & 0 \\*
\hspace{0.7em}Astra, r1 & 75.0 & 45.0 & 71.9 & 50.0 & 39.3 & 25.0 & 0.0 & 10.0 & 42.5 & 42.5 & 36.0 & 0 \\*
\hspace{0.7em}Astra, r2 & 70.8 & 30.0 & 65.6 & 45.8 & 42.9 & 27.5 & 0.0 & 20.0 & 40.0 & 45.0 & 36.0 & 0 \\
\end{longtable}}

%% file: tables/tabA_t1_fits_r1.tex
{\setlength{\tabcolsep}{1.5pt}%
\begin{tabularx}{\linewidth}{@{}l*{10}{>{\centering\arraybackslash}X}>{\centering\arraybackslash}p{4.2em}>{\centering\arraybackslash}p{3.4em}@{}}
\toprule
& \multicolumn{4}{c}{\textbf{Design \& planning}} & \multicolumn{4}{c}{\textbf{Experimental execution}} & \multicolumn{2}{c}{\textbf{Analysis \& writing}} & & \\
\cmidrule(lr){2-5}\cmidrule(lr){6-9}\cmidrule(lr){10-11}
\multicolumn{1}{c}{\textbf{System}} & \scriptsize Lit. & \scriptsize Claims & \scriptsize Meth. & \scriptsize Plan & \scriptsize Module & \scriptsize Code & \scriptsize Result & \scriptsize Claim & \scriptsize Anal. & \scriptsize Writ. & \scriptsize\textbf{Overview} & \scriptsize\textbf{Claims met} \\
\midrule
\multicolumn{13}{@{}l}{\textbf{General-purpose agent harness}}\\
\hspace{0.7em}Codex / Sol & 100.0 & 63.6 & 100.0 & 30.8 & 75.0 & 38.6 & 100.0 & 100.0 & 59.1 & 84.4 & 75.6 & 11/11 \\
\hspace{0.7em}Codex / Astra & 50.0 & 36.4 & 100.0 & 98.1 & 75.0 & 36.4 & 0.0 & 4.5 & 50.0 & 90.6 & 52.4 & 0/11 \\
\hspace{0.7em}Claude Code / Opus 4.8 & 94.4 & 0.0 & 62.5 & 46.2 & 85.0 & 34.1 & 4.5 & 4.5 & 0.0 & 33.3 & 35.5 & 0/11 \\
\hspace{0.7em}Kimi Code / K3 & 100.0 & 100.0 & 100.0 & 36.5 & 70.0 & 43.2 & 31.8 & 47.7 & 31.8 & 87.5 & 61.4 & 2/11 \\
\multicolumn{13}{@{}l}{\textbf{Research-orchestration system (Sol)}}\\
\hspace{0.7em}ARIS / Sol & 100.0 & 77.3 & 100.0 & 38.5 & 75.0 & 43.2 & 0.0 & 18.2 & 56.8 & 93.8 & 55.9 & 0/11 \\
\hspace{0.7em}Orchestra / Sol & 100.0 & 97.7 & 100.0 & 40.4 & 100.0 & 84.1 & 72.7 & 72.7 & 36.4 & 87.5 & 79.1 & 8/11 \\
\hspace{0.7em}AutoResearchClaw / Sol & 0.0 & 13.6 & 100.0 & 0.0 & 45.0 & 4.5 & 0.0 & 0.0 & 9.1 & 51.0 & 22.9 & 0/11 \\
\hspace{0.7em}DeepScientist / Sol & 100.0 & 0.0 & 100.0 & 42.3 & 0.0 & 0.0 & 36.4 & 36.4 & 0.0 & 84.4 & 38.0 & 4/11 \\
\midrule
\textbf{Mean of the eight} & 80.6 & 48.6 & 95.3 & 41.6 & 65.6 & 35.5 & 30.7 & 35.5 & 30.4 & 76.6 & 52.6 & 25/88 \\
\midrule
\multicolumn{13}{@{}l}{\textbf{Stagewise runs of the three harnesses run in both modes}}\\
\hspace{0.7em}Codex / Sol & 88.9 & 31.8 & 100.0 & 98.1 & 100.0 & 88.6 & 88.6 & 88.6 & 63.6 & 72.9 & 84.6 & 9/11 \\
\hspace{0.7em}Claude Code / Opus 4.8 & 63.9 & 29.5 & 100.0 & 73.1 & 100.0 & 47.7 & 50.0 & 50.0 & 84.1 & 84.4 & 69.0 & 5/11 \\
\hspace{0.7em}Kimi Code / K3 & 100.0 & 100.0 & 100.0 & 94.2 & 100.0 & 97.7 & 97.7 & 97.7 & 61.4 & 93.8 & 94.8 & 10/11 \\
\bottomrule
\end{tabularx}}